\documentclass{article}
\pdfoutput=1

\PassOptionsToPackage{round,authoryear}{natbib}
\usepackage[preprint]{neurips_2026}

\usepackage[utf8]{inputenc}
\usepackage[T1]{fontenc}
\usepackage{hyperref}
\hypersetup{
  pdftitle={Rendering-Aware Bayesian 3D Gaussian Splatting with Native Uncertainty and Adaptive Complexity Control},
  pdfauthor={Gaoxiang Jia, Vikram Appia, Junzhou Huang, Xinlei Wang},
  hypertexnames=false
}
\usepackage{url}
\usepackage{booktabs}
\usepackage{amsfonts}
\usepackage{amsmath}
\usepackage{amssymb}
\usepackage{amsthm}
\usepackage{graphicx}
\usepackage{multirow}
\usepackage{nicefrac}
\usepackage{microtype}
\usepackage{xcolor}
\usepackage{enumitem}
\usepackage{algorithm}
\usepackage{algorithmic}

\newcommand{\MethodStd}{Standard 3DGS}
\newcommand{\MethodNIW}{Bayesian 3DGS-NIW}
\newcommand{\MethodOfficial}{Official 3DGS Reproduction}

\newcommand{\MethodDPMM}{DPMM-3DGS}

\title{Rendering-Aware Bayesian 3D Gaussian Splatting with Native Uncertainty and Adaptive Complexity Control}

\author{
  Gaoxiang Jia\thanks{Co-corresponding authors.} \\
  Advanced Micro Devices, Inc.\\
  \texttt{jiagaoxiang@gmail.com} \\
  \And
  Vikram Appia \\
  Advanced Micro Devices, Inc.\\
  \AND
  Junzhou Huang \\
  Dept.\ of Computer Science and Engineering\\
  University of Texas at Arlington\\
  \And
  Xinlei Wang\footnotemark[1] \\
  Dept.\ of Mathematics \&\\Division of Data Science\\
  University of Texas at Arlington\\
  \texttt{xinlei.wang@uta.edu} \\
}

\begin{document}

\maketitle

\begin{abstract}
3D Gaussian splatting (3DGS) is a strong representation for real-time, high-fidelity novel-view synthesis, yet its standard training pipeline relies on point estimates and hand-tuned heuristics that do not natively provide uncertainty or principled complexity control. This limitation is most apparent under limited view support or fixed acquisition budgets, where a point estimate cannot identify weakly supported geometry or rank informative candidate views. We introduce a rendering-aware Bayesian framework that tracks 3DGS geometry with a Normal-Inverse-Wishart (NIW) posterior over Gaussian means and covariances using renderer-derived surrogate summaries; an optional Dirichlet-process extension turns component usage into a probabilistic complexity signal, and the training schedule makes the closed-form-versus-approximate inference boundary explicit. Re-rendering posterior geometry samples produces native predictive uncertainty for interval calibration and active view selection. In a fixed-budget 16-to-32 active-view-selection task, native NIW acquisition improves PSNR by $+0.453$~dB and LPIPS by $-0.0146$ over a scoring-only 3-member standard-ensemble acquisition baseline, winning 29/39 scene-seed pairs and 10/13 scene means on PSNR, with PSNR gains also over PPU-style ($+0.355$~dB) and NIW proxy ($+0.401$~dB) acquisition. NIW native intervals reduce 95\% coverage error by about $17\times$ relative to a shared proxy ($|\mathrm{Cov@95}-0.95|$: $0.046$ versus $0.796$) and are about $10\times$ closer to nominal coverage than a 3-member deep ensemble ($0.047$ versus $0.454$) at roughly one-third the training cost. As a reconstruction compatibility check, paired NIW-vs-standard analysis over 39 scene-seed runs yields $+0.030$~dB PSNR (95\% cluster bootstrap CI $[{+}0.004, {+}0.059]$~dB) for 1.6\% additional training time. Together, these results position Bayesian 3DGS as a practical probabilistic scene representation for decision-facing tasks such as active view selection.

\end{abstract}

\section{Introduction}
\label{sec:introduction}

\begin{figure*}[t]
  \centering
  \includegraphics[width=\linewidth]{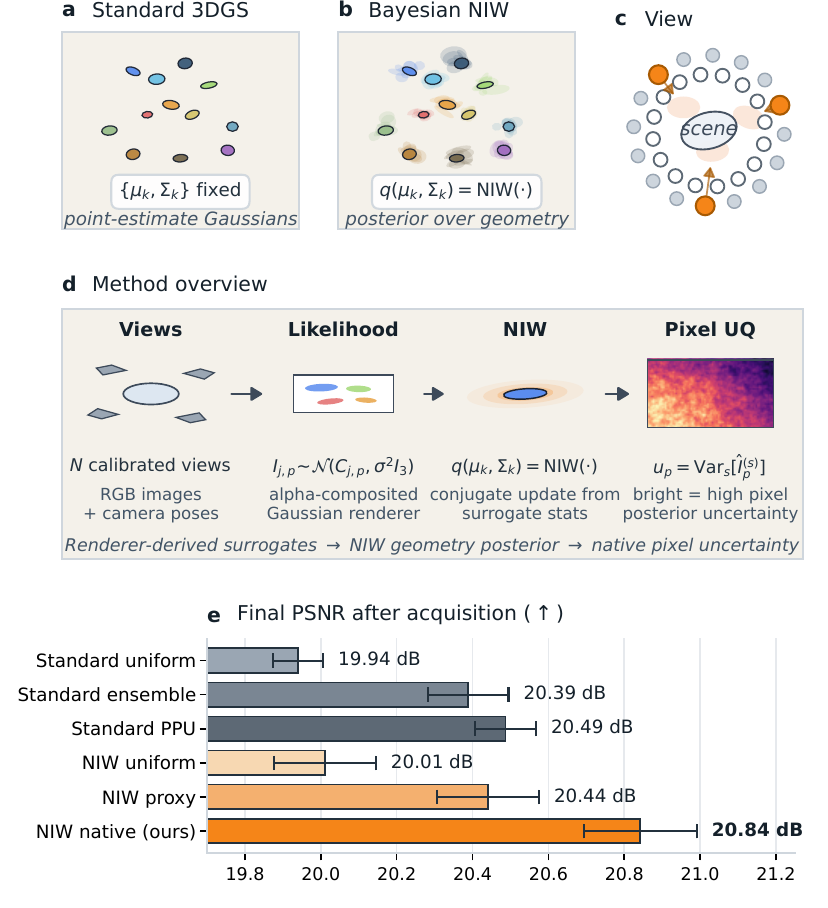}
  \caption{Rendering-aware Bayesian 3DGS. (a) Standard 3DGS keeps $\{\mu_k,\Sigma_k\}$ as point estimates. (b) Bayesian 3DGS-NIW attaches a posterior $q(\mu_k,\Sigma_k)$; translucent clouds show posterior geometry draws, with tighter clouds for better-supported components. (c) In fixed-budget $16\!\rightarrow\!32$ active-view selection, native uncertainty selects views near high-variance regions. (d) Camera-calibrated views define a rendering likelihood, renderer-derived surrogate statistics feed the NIW update, and posterior re-rendering yields pixel uncertainty. (e) Final PSNR reaches $20.84$~dB, $+0.453$~dB over a scoring-only standard-ensemble baseline. Error bars show $\pm$1 standard error of the mean (SEM) across three seeds, averaged over 13 scenes.}
  \label{fig:teaser}
\end{figure*}

3D Gaussian splatting (3DGS) has become a central representation for novel-view synthesis, combining strong reconstruction quality with real-time rendering \citep{kerbl2023gaussiansplatting}. Yet the standard pipeline provides no posterior characterization of scene geometry: the representation is a point estimate, Gaussian birth and death are governed by hand-tuned gradient and opacity thresholds, and the final model offers no native uncertainty characterization.

This absence is least visible in the fully trained canonical benchmark, where a strong point-estimate 3DGS model is already difficult to surpass by large margins. It becomes a clear failure mode when 3DGS must act under limited support or fixed budgets: a point estimate cannot natively rank informative candidate views or identify weakly supported geometry. We therefore ask which parts of the geometry carry high posterior variance, how that uncertainty propagates to rendered pixels, and whether it improves a downstream decision such as fixed-budget view acquisition. Recent work has advanced probabilistic optimization, uncertainty estimation, and pruning for Gaussian splatting \citep{kheradmand2024mcmc,li2024multiscaleuq,savant2024modeling,vandemaele2025vbgs,hanson2025pup,wu2026horseshoe,galappaththige2026photometric}; our focus is the less-explored combination of closed-form geometry updates from renderer-derived surrogate summaries, native geometry-induced predictive uncertainty, and a Dirichlet-process (DP) complexity prior in one auditable 3DGS runtime. Appendix~\ref{sec:related} provides an extended discussion.

We address these questions by embedding the 3DGS renderer inside a probabilistic observation model. Our primary formulation places a normal-inverse-Wishart (NIW) posterior over each Gaussian's mean and covariance, updated online via renderer-derived surrogate summaries that play the role of Gaussian complete-data statistics (Figure~\ref{fig:teaser}a,b,d). The same alpha-compositing renderer that produces colors at training time also propagates this geometry posterior to per-pixel predictive variance, supplying an interval-calibration signal and, when averaged over a candidate camera, an acquisition heuristic for fixed-budget view selection. An optional DP layer models component usage probabilistically, providing complexity signals that complement standard densification. A precise caveat: ``rendering-aware'' means that the observation model and surrogate summaries are derived from the alpha-composited renderer; it does not mean exact end-to-end Bayesian inference through the full nonlinear rendering pipeline. The inference schedule therefore combines closed-form updates within stated submodels with explicitly approximate steps, and Table~\ref{tab:inference_boundary} makes the boundary visible.

The clearest practical signal appears in a fixed-budget 16-to-32 active-view-selection task: native NIW acquisition uses posterior uncertainty to choose cameras near high-variance geometry (Figure~\ref{fig:teaser}c) and improves final quality by $+0.453$~dB and $-0.0146$ LPIPS over a scoring-only standard-ensemble acquisition baseline (Figure~\ref{fig:teaser}e), winning 29/39 scene-seed runs and 10/13 scene means on PSNR. The native NIW posterior also provides substantially improved interval calibration, achieving a $17\times$ reduction in coverage error relative to the shared proxy ($0.046$ versus $0.796$). Against a 3-member deep ensemble, NIW is not the strongest ranking-style uncertainty baseline, but it is about $10\times$ closer to nominal interval coverage ($0.047$ versus $0.454$) while avoiding the ensemble's roughly $3\times$ training cost (2413~s versus 815~s). On the canonical 13-scene benchmark, \MethodNIW{} shows a small paired PSNR gain of $+0.030$~dB over \MethodStd{} (95\% cluster bootstrap CI $[{+}0.004, {+}0.059]$~dB) with 1.6\% training-time overhead, indicating compatibility rather than a primary source of improvement. The DP extension similarly provides a secondary compactness operating point rather than competing with dedicated compression systems \citep{hanson2025pup,fan2024lightgaussian,lee2024compact3dgs}. Overall, the contribution is a probabilistic scene representation with native predictive uncertainty, transparent approximation boundaries, and data-driven complexity control over a strong rendering backbone.

\paragraph{Relation to existing work.}
The closest probabilistic 3DGS methods take either a stochastic-optimization view or a detached variational view: 3DGS-MCMC~\citep{kheradmand2024mcmc} uses stochastic-gradient Langevin dynamics, and VBGS~\citep{vandemaele2025vbgs} maintains conjugate posteriors on a density model detached from the renderer. Uncertainty-aware methods add predictive heads or sample-based surrogates~\citep{savant2024modeling,li2024multiscaleuq,galappaththige2026photometric}, while pruning methods pursue post-hoc sparsity~\citep{hanson2025pup,wu2026horseshoe}. Our formulation differs by updating an NIW posterior from renderer-derived surrogate summaries in a single training trajectory and coupling it to an explicit nonparametric complexity prior. Deep ensembles~\citep{lakshminarayanan2017deepensembles} capture a different posterior object at higher cost, so we report both regimes side by side.

Our contributions are organized around why Bayesian structure should help, and where our evidence is new. The general premise is not new: Bayesian experimental design and active learning motivate uncertainty-guided measurement selection \citep{mackay1992information,settles2009active}, while conjugate Bayesian updates and Bayesian nonparametric priors are standard tools for probabilistic updating and adaptive model complexity \citep{murphy2012mlapp,ferguson1973bayesian,sethuraman1994constructive,blei2006variationaldp}. Our contribution is to make those ideas rendering-aware, computationally practical for 3DGS, and empirically verified under matched 3DGS protocols:
\begin{enumerate}[leftmargin=1.5em]
  \item \textbf{Rendering-aware Bayesian 3DGS.} We introduce an NIW geometry posterior whose dispersion adapts to renderer-derived support, together with a DP extension that provides a probabilistic usage signal. We explicitly delineate the boundary between closed-form and approximate inference, clarifying which components support Bayesian interpretation.
  \item \textbf{Decision-facing validation.} We show that geometry-induced predictive uncertainty can drive an acquisition score for fixed-budget active view selection, and evaluate it directly against ensemble and predictive photometric uncertainty (PPU)-style acquisition baselines \citep{galappaththige2026photometric} rather than claiming optimal Bayesian experimental design.
  \item \textbf{Component-level empirical analysis.} We isolate the contribution of each module: NIW preserves the rendering backbone, native posterior sampling improves interval calibration and acquisition, and the DP layer provides a compactness operating point, while reconstruction, compression, and out-of-distribution (OOD) detection remain explicitly bounded claims.
\end{enumerate}

\section{Method}
\label{sec:method}

\subsection{Rendering-aware Bayesian 3DGS}
\label{subsec:niw-model}

Our starting point is a probabilistic reinterpretation of 3DGS in which posterior structure is placed on the same Gaussian geometry that the renderer projects to pixels. Let $\Theta=\{(\mu_k,\Sigma_k,o_k,c_k)\}_{k=1}^K$ be the per-Gaussian parameters and $a\in\{0,1\}^K$ the component activities. The standard 3DGS splat renderer produces color $C_{j,p}(\Theta,a)$ at pixel $p$ in camera $j$ by depth-ordered alpha compositing of visible Gaussians, and we model the observation as $I_{j,p} \sim \mathcal{N}(C_{j,p}(\Theta,a), \sigma^2 I_3)$. We place a normal-inverse-Wishart (NIW) prior on each $(\mu_k,\Sigma_k)$ pair (Eq.~\ref{eq:niw-prior}) and an optional truncated Dirichlet process activity prior (Eq.~\ref{eq:stick-breaking}). The factors are familiar Bayesian objects---a Gaussian pixel likelihood, the canonical conjugate prior for a Gaussian likelihood with unknown mean and covariance, and a standard nonparametric prior for adaptive component count \citep{ferguson1973bayesian,sethuraman1994constructive,blei2006variationaldp}---arranged so that the renderer mediates between geometry and image (Figure~\ref{fig:teaser}d).

We call the formulation \emph{rendering-aware} in the specific sense that the NIW posterior is updated from sufficient-statistic surrogates $(\hat{N}_k,\bar{x}_k,C_k)$ computed from the alpha-compositing renderer's own state, rather than from photometric residuals or a detached density model. Operationally, the renderer induces for each component an auxiliary spatial data set represented only by these summaries; under the surrogate model $z_{k,i}\sim\mathcal{N}(\mu_k,\Sigma_k)$, the NIW update is closed-form. The pseudo-observations $z_{k,i}$ are not observed 3D points, so the approximation is concentrated in how rendering state is summarized, not in the conjugate algebra. We therefore make no claim of exact end-to-end Bayes through the full nonlinear renderer, and Table~\ref{tab:inference_boundary} marks which factors are closed-form and which remain approximate. The live trainer still optimizes the standard photometric / SSIM rendering loss, with this likelihood supplying the probabilistic block used for posterior tracking.

The NIW prior on each Gaussian is
\begin{equation}
(\mu_k, \Sigma_k) \sim \operatorname{NIW}(m_0, \kappa_0, \nu_0, S_0),
\label{eq:niw-prior}
\end{equation}
which yields a natural Bayesian object for online geometric uncertainty \citep{murphy2012mlapp}. We denote posterior factors by $q(\cdot)$ throughout, including factors with closed-form updates under the stated surrogate model. We use a weakly informative prior centred on the initial point cloud: $\nu_0 = D + 2 = 5$ is the minimally informative degrees-of-freedom that keeps the inverse-Wishart proper \citep{murphy2012mlapp}, and $\kappa_0 = 0.01$ leaves the mean nearly flat, so rendering evidence quickly dominates. The conjugate update requires an effective count $\hat{N}_k$, weighted mean $\bar{x}_k$, and scatter $C_k$. In a classical mixture model these come from responsibilities; here they are derived from renderer state. In the benchmarked NIW rows, $\hat{N}_k$ is an exponential moving average (EMA)-smoothed opacity-times-diagonal-volume proxy, $\hat{N}_k \propto o_k \sqrt{\prod_d [\mathbb{E}[\Sigma_k]]_{dd}}$, normalized to the effective pixel count, while the pseudo-observation is the current Gaussian center and the online update sets $C_k=0$.
\begin{align}
\kappa_n &= \kappa_0 + \hat{N}_k, \qquad
m_n = \frac{\kappa_0 m_0 + \hat{N}_k \bar{x}_k}{\kappa_n}, \label{eq:niw-update-1} \\
\nu_n &= \nu_0 + \hat{N}_k, \qquad
S_n = S_0 + C_k + \frac{\kappa_0 \hat{N}_k}{\kappa_n}(\bar{x}_k - m_0)(\bar{x}_k - m_0)^\top. \label{eq:niw-update-2}
\end{align}
The posterior expected covariance is $\mathbb{E}[\Sigma_k] = S_n / (\nu_n - D - 1)$ and the posterior mean variance is $\operatorname{Var}[\mu_k] = \mathbb{E}[\Sigma_k] / \kappa_n$, where $D=3$. Because $\hat{N}_k$ grows with rendering support, well-observed components concentrate while occluded or under-observed components retain high posterior variance, giving an auditable rendering-aware geometry posterior with the approximation boundary stated in Table~\ref{tab:inference_boundary}.

\subsection{Native posterior uncertainty}
\label{subsec:native-uq}

The NIW posterior enables a native source of geometry-induced predictive uncertainty not provided by the matched point-estimate control. Given the learned posterior $q(\mu_k, \Sigma_k)$, we draw $S$ posterior samples and re-render the scene:
\begin{equation}
\hat{I}^{(s)}_{j,p} = C_{j,p}(\Theta^{(s)}, a), \quad \Theta^{(s)} \sim q(\Theta), \quad
u_p = \frac{1}{S}\sum_{s=1}^{S}\left\|\hat{I}^{(s)}_{j,p} - \bar{I}_{j,p}\right\|^2,
\label{eq:native-uq}
\end{equation}
where $\bar{I}_{j,p}$ is the sample mean and $u_p$ is the per-pixel uncertainty. In the NIW-only headline native uncertainty quantification (UQ) rows, $\Theta^{(s)}$ samples only the geometry posterior; opacity and color remain fixed, and aleatoric variance enters through a plug-in residual mean squared error (MSE) estimate $\hat{\sigma}^2$. The fuller Dirichlet process mixture model (DPMM) predictive path also samples the inverse-gamma noise posterior and approximate activity, opacity, and color posterior draws; Table~\ref{tab:inference_boundary} marks this split. The publication benchmark uses $S{=}50$ samples for both shared proxy-UQ and Bayesian native-posterior-UQ estimates, counted as evaluation rather than training cost.

We distinguish this native geometry uncertainty from the shared proxy protocol used for fair cross-method comparison. The proxy protocol applies identical position-jitter Monte Carlo to all methods---including deterministic baselines---so that ranking-style calibration metrics, including area under the sparsification error (AUSE) and expected calibration error (ECE), remain comparable; it answers the \emph{fairness} question. Native posterior uncertainty answers the \emph{capability} question: whether the Bayesian model can express meaningful uncertainty from its own learned posterior, without external perturbation. For active view selection, a candidate camera receives the mean pixel score $\frac{1}{|\mathcal{P}|}\sum_{p\in\mathcal{P}} u_p$ and we add the top 16. This is a variance-based acquisition heuristic: large geometry-induced dispersion at a candidate view is a proxy for where the model is unsure \citep{mackay1992information,settles2009active}, not a claim to optimize mutual information under nonlinear splatting.

\subsection{Dirichlet process complexity extension}
\label{subsec:dp-extension}

To model component usage probabilistically, we add a truncated stick-breaking construction,
\begin{equation}
v_k \sim \operatorname{Beta}(1, \alpha), \qquad
\pi_k = v_k \prod_{\ell < k} (1 - v_\ell),
\label{eq:stick-breaking}
\end{equation}
and optimize its variational form with coordinate-ascent variational inference \citep{ferguson1973bayesian,sethuraman1994constructive,blei2006variationaldp}. Gaussians whose posterior mixing weight $\mathbb{E}[\pi_k]$ falls below a threshold are pruned, while $\alpha$ controls consolidation. We set $\alpha = 5$ by default; with $K \sim 10^6$, the prior expected active count is approximately $\alpha \log K \approx 69$ \citep{blei2006variationaldp}, encouraging consolidation without extreme sparsity. In the current trainer, this DP machinery augments rather than replaces standard 3DGS densification and housekeeping.

This DPMM extension creates a second operating point with fewer retained Gaussians, faster rendering, and posterior summaries reusable for fixed-budget compression. It complements rather than replaces NIW: the main quality story is carried by the NIW posterior, while the DP layer provides adaptive complexity control.

\subsection{Hybrid inference and approximation transparency}
\label{subsec:hybrid-inference}

Our live GPU trainer is a staged hybrid Bayesian runtime rather than a literal end-to-end evidence lower bound (ELBO) optimizer. It combines closed-form submodel updates and explicitly approximate updates around the standard 3DGS optimization path, staged for stability. The corresponding staged training procedure is given in Algorithm~\ref{alg:training} in Appendix~\ref{sec:supp-boundary}. Table~\ref{tab:inference_boundary} delineates the boundary: geometry ($\mu_k, \Sigma_k$), stick-breaking ($v_k$), and noise ($\sigma^2$) carry the clearest explicit posterior structure, while opacity and color receive slower approximate posterior refinements around the standard optimizer.

\begin{table}[t]
  \caption{Closed-form-versus-approximate boundary for the main update blocks. The NIW spatial posterior and stick-breaking weights have closed-form updates within the stated surrogate or variational submodel, while opacity and color remain approximate refinements around the standard optimizer.}
  \label{tab:inference_boundary}
  \centering
  \footnotesize
  \begin{tabular}{p{0.25\linewidth}p{0.22\linewidth}p{0.39\linewidth}}
    \toprule
    Block & Status & Role \\
    \midrule
    Stick-breaking $q(v_k)$ & Closed-form coordinate-ascent variational inference (CAVI) & Beta update for DP complexity given factorization \\
    NIW spatial $q(\mu_k,\Sigma_k)$ & Conjugate given summaries & Core Bayesian geometry block \\
    Effective counts $\hat{N}_k$ & Approximate & Rendering-derived proxy for component support \\
    Activity $q(a_k)$ & Approximate & Fast usage gate for complexity control \\
    Opacity $o_k$ & Approximate posterior refinement & Opacity block refined around the standard optimizer \\
    Color $c_k$ & Approximate posterior refinement & Appearance block refined around the standard optimizer \\
    Noise variance $q(\sigma^2)$ & Closed-form inverse-gamma (IG) step & Given minibatch residual statistics in the fuller DPMM path \\
    \bottomrule
  \end{tabular}
\end{table}

Training is staged around the standard 3DGS optimizer so the point-estimate rendering state remains stable while posterior blocks accumulate evidence. In the 30k canonical DPMM row, NIW geometry tracking is active from the start, stick-breaking CAVI runs every 50 iterations, activity and residual-noise updates begin after 5k iterations, and slower opacity/color refinements begin after 15k; Appendix~\ref{subsec:supp-canonical} records the full schedule. The native-posterior calibration results in Section~\ref{subsec:calibration} are driven by the closed-form NIW geometry block; the fuller DPMM predictive path also uses closed-form inverse-gamma noise updates, while activity, opacity, and color remain explicitly approximate. The resulting system provides a rendering-aware posterior over geometry and complexity with the boundary stated in Table~\ref{tab:inference_boundary}.

\section{Experiments}
\label{sec:experiments}

\subsection{Protocol and baselines}
\label{subsec:protocol}

We evaluate on the standard 13-scene suite spanning Mip-NeRF~360 \citep{barron2022mipnerf360}, Tanks and Temples \citep{knapitsch2017tanksandtemples}, and Deep Blending \citep{hedman2018deepblending}. The canonical table contains methods rerun end-to-end under one held-out protocol: \MethodStd{} (matched deterministic control), \MethodNIW{} (primary Bayesian model), \MethodDPMM{} (compactness extension), \MethodOfficial{} (external 3DGS anchor), and 3DGS-MCMC \citep{kheradmand2024mcmc}. Stronger uncertainty and pruning comparators enter as focused follow-ups when their native objective matches the study, so deep ensembles and predictive photometric uncertainty (PPU) are used for uncertainty quantification (UQ)/acquisition while principled uncertainty pruning (PUP) is scoped to compression. Because the primary claim is representation capability rather than a large reconstruction gain, the canonical benchmark is a compatibility check, complemented by a paired NIW-vs-standard analysis over matched scene-seed runs.

We report peak signal-to-noise ratio (PSNR), structural similarity index measure (SSIM), and learned perceptual image patch similarity (LPIPS) \citep{zhang2018lpips} for quality; retained Gaussian count, frames per second (FPS), and train time for compactness--efficiency; and uncertainty metrics under two regimes. The fair all-method comparison uses a shared position-jitter Monte Carlo proxy with area under the sparsification error (AUSE) and expected calibration error (ECE). Bayesian rows additionally expose native geometry-induced predictive uncertainty, where we emphasize likelihood-based calibration and $|\mathrm{Cov@95}-0.95|$. Thus the active-view study tests decision value, native-UQ tests interval calibration, the canonical benchmark tests compatibility, and compression remains a bounded matched-checkpoint follow-up.

\subsection{Downstream task: active view selection}
\label{subsec:active_view}

The most concrete failure mode studied in this paper is decision-making under limited view support. When only 16 train views are initially available and the budget permits 16 more, a point-estimate 3DGS model must still answer ``which view should we acquire next?'' using deterministic surrogates such as uniform spacing, disagreement heuristics, or post-hoc uncertainty scores rather than a native posterior over unsupported geometry. We therefore evaluate a fixed-budget active-view task that converts uncertainty quality into a rendering outcome: under the same acquisition and training budget, a useful uncertainty signal should choose views that improve the final held-out reconstruction, not merely score pixels after the fact. For each scene, we begin from the same deterministic 16-view seed subset, train a seed model, rank remaining train views by the policy score, add the top 16 views, and continue training to the same 30k-iteration budget. Scored policies use mean candidate-view uncertainty: cross-member red-green-blue (RGB) variance for the scoring-only ensemble, residual-head uncertainty for the PPU-style follow-up \citep{galappaththige2026photometric}, shared proxy uncertainty for NIW proxy, and Eq.~\ref{eq:native-uq} for NIW native. The comparison family is fixed rather than chosen scene-by-scene: every scene and seed in the five-policy core uses the same 16-to-32 acquisition ladder, and Table~\ref{tab:active_view_selection} adds the protocol-matched PPU follow-up on frozen standard checkpoints.

\begin{table}[t]
  \caption{Active view selection (13 scenes, 3 seeds). Each policy starts from the same 16-view seed subset, selects 16 additional views, and trains to the same 30k-iteration budget. The ensemble row uses three standard phase-1 models for scoring only; the PPU row is a frozen-checkpoint follow-up. Native NIW acquisition gives the strongest final quality.}
  \label{tab:active_view_selection}
  \centering
  \footnotesize
  \begin{tabular}{lcccc}
    \toprule
    Policy & PSNR$\uparrow$ & SSIM$\uparrow$ & LPIPS$\downarrow$ & Time (s)$\downarrow$ \\
    \midrule
    Standard + uniform add & 19.940 & 0.6307 & 0.3485 & \textbf{873.9} \\
    Standard + ensemble-UQ add & 20.389 & 0.6367 & 0.3428 & 1531.2 \\
    Standard + PPU-style add & 20.487 & 0.6457 & 0.3376 & 960.9 \\
    NIW + uniform add & 20.011 & 0.6305 & 0.3486 & 1009.2 \\
    NIW + proxy-UQ add & 20.441 & 0.6519 & 0.3341 & 888.7 \\
    NIW + native-UQ add & \textbf{20.842} & \textbf{0.6577} & \textbf{0.3282} & 1233.1 \\
    \bottomrule
  \end{tabular}
\end{table}

The downstream conclusion remains positive under stronger baselines. The scoring-only ensemble rule is nontrivial, improving over uniform standard acquisition by $+0.449$~dB, and the protocol-matched PPU-style row improves by a further $+0.0979$~dB. Even so, native NIW acquisition remains strongest: relative to the ensemble row, it improves PSNR by $+0.453$~dB and LPIPS by $-0.0146$, with 29/39 PSNR wins and 10/13 winning scene means; relative to the PPU-style row, it improves PSNR by $+0.355$~dB. The selected views are also genuinely different (mean Jaccard overlap 0.056 versus NIW proxy and 0.059 versus standard ensemble). Native NIW is not uniformly best---the Deep Blending indoor scenes favor ensemble acquisition---and it costs 39\% more than NIW proxy acquisition, but it is still 19\% faster than scoring-only standard ensembles. This is the clearest practical result: native Bayesian uncertainty changes which cameras are acquired and improves final reconstruction against generic and 3DGS-specific selectors.

Appendix Table~\ref{tab:component_evidence} provides a component-level summary that separates the intended role of each block. We keep the main text focused on the direct active-view, calibration, and compatibility evidence rather than treating all metrics as one leaderboard.

\subsection{Native posterior uncertainty}
\label{subsec:calibration}

The central Bayesian contribution is the native geometry-induced predictive uncertainty, which the matched deterministic control does not natively provide. We therefore compare two uncertainty regimes within each Bayesian row: the shared proxy for fair cross-method ranking and the learned posterior for Bayesian capability. Here \(\mathrm{Cov@95}\) is the empirical fraction of held-out pixel residuals inside nominal Gaussian 95\% intervals constructed from each uncertainty recipe. Within this Bayesian-only comparison, the NIW native posterior is slightly conservative but about $17\times$ closer to nominal coverage than the shared proxy ($|\mathrm{Cov@95} - 0.95|$: $0.046$ versus $0.796$), and the same shift is even larger for DPMM ($0.026$ versus $0.782$). In plain terms, nominal 95\% NIW intervals contain $\approx\!99.6\%$ of held-out pixels (modestly conservative), whereas shared-proxy intervals contain only $\approx\!15.4\%$ (heavily under-covering); a calibrated interval is what a downstream consumer needs when asking which rendered-pixel error bar to trust. Under the likelihood scoring convention for each Bayesian row, native posterior NLL is lower than the proxy on every scene; because proxy and native variances are constructed differently, we use NLL as supporting evidence rather than as a cross-method ranking. The tradeoff is that the shared proxy remains stronger on ranking-style metrics: native posterior AUSE is worse on 13/13 scenes for both NIW and DPMM, and native ECE is worse on all 13 NIW scenes and all but 2 DPMM scenes. The mechanism is intuitive: position-jitter Monte Carlo orders pixels by local rendering sensitivity (helpful for AUSE/ECE) but produces intervals far too narrow for absolute coverage, while native posterior intervals are calibrated for coverage but less aligned with per-pixel residual magnitude. We therefore use the proxy protocol for fair all-method ranking and the native posterior for the interval-calibration claim.

As a posterior-behavior diagnostic, per-scene mean pixel uncertainty correlates with reconstruction difficulty ($-$PSNR), with Spearman $\rho = 0.73$ for NIW and $\rho = 0.90$ for DPMM over the 13 scenes. The supplement reports the per-scene tables, uncertainty magnitudes, and an outlier-transparent sensitivity analysis for the NIW--LPIPS trend.

\paragraph{External UQ baseline: deep ensembles.}
A natural concern is whether the native-posterior story only beats a weak proxy baseline. We therefore also evaluate a 3-member deep ensemble from independently trained \MethodStd{} runs on the same scene suite. Table~\ref{tab:uq_baseline_comparison} should be read as a trade: the ensemble is stronger on reconstruction and ranking-style uncertainty, whereas NIW native is about $10\times$ closer to nominal interval coverage ($0.047$ versus $0.454$) while costing 815~s rather than 2413~s. The matched \MethodStd{} and \MethodNIW{} rows come from the internal-checkpoint replay path used for the ensemble, so their third-decimal values differ slightly from Table~\ref{tab:main_results}. The rows quantify different posterior objects---cross-member predictive disagreement versus explicit geometry uncertainty inside one trained representation---so NIW is a single-trajectory Bayesian representation with strong interval calibration, not a universal winner on every uncertainty metric. We report negative log-likelihood (NLL) only within a fixed uncertainty construction rather than as a cross-regime ranking.

\begin{table}[t]
  \caption{Finished UQ comparison on the same 13-scene suite, evaluated through the internal-checkpoint replay path used for the ensemble baseline. Shared proxy, native posterior, and deep-ensemble uncertainty are different predictive regimes; the table separates ranking-style uncertainty from interval calibration. Ensemble time sums all three members; NIW rows report one trajectory.}
  \label{tab:uq_baseline_comparison}
  \centering
  \scriptsize
  \setlength{\tabcolsep}{3.5pt}
  \begin{tabular}{lcccccc}
    \toprule
    Row & PSNR$\uparrow$ & LPIPS$\downarrow$ & AUSE$\downarrow$ & ECE$\downarrow$ & $|\mathrm{Cov@95}{-}0.95|$$\downarrow$ & Time (s)$\downarrow$ \\
    \midrule
    \MethodStd{} (shared proxy) & 27.322 & 0.2113 & 0.0214 & 0.0457 & 0.795 & \textbf{804} \\
    \MethodNIW{} (shared proxy) & 27.332 & 0.2111 & 0.0213 & 0.0450 & 0.795 & 815 \\
    \MethodNIW{} (native posterior) & 27.332 & 0.2111 & 0.0288 & 0.5991 & \textbf{0.047} & 815 \\
    Deep ensemble (3 members) & \textbf{27.895} & \textbf{0.2070} & \textbf{0.0191} & \textbf{0.0277} & 0.454 & 2413 \\
    \bottomrule
  \end{tabular}
\end{table}

\subsection{Canonical benchmark results}
\label{subsec:canonical}

Table~\ref{tab:main_results} presents the canonical compatibility benchmark, not the paper's sole practical-value evidence. \MethodNIW{} reaches 27.333 PSNR versus 27.303 for the matched deterministic control and 27.304 for \MethodOfficial{}, with the lowest LPIPS among the five rows. The table shows that Bayesian structure preserves competitive rendering quality.

\begin{table}[t]
  \caption{Canonical 13-scene compatibility benchmark (3-seed macro averages). All methods use shared proxy-UQ for AUSE and ECE. NIW has the highest macro PSNR and lowest LPIPS; DPMM trades quality for 36\% fewer Gaussians and 44\% faster rendering.}
  \label{tab:main_results}
  \centering
  \scriptsize
  \setlength{\tabcolsep}{3.5pt}
  \begin{tabular}{lcccccccc}
    \toprule
    Method & PSNR$\uparrow$ & SSIM$\uparrow$ & LPIPS$\downarrow$ & AUSE$\downarrow$ & ECE$\downarrow$ & Gauss. & FPS$\uparrow$ & Time (s)$\downarrow$ \\
    \midrule
    \MethodStd{} & 27.303 & 0.8358 & 0.2113 & \textbf{0.0213} & 0.0452 & 2.532M & 267.3 & \textbf{802.3} \\
    \MethodNIW{} & \textbf{27.333} & 0.8359 & \textbf{0.2111} & 0.0213 & \textbf{0.0452} & 2.547M & 267.2 & 815.1 \\
    \MethodOfficial{} & 27.304 & \textbf{0.8359} & 0.2113 & 0.0241 & 0.0591 & 2.519M & 279.4 & 815.0 \\
    3DGS-MCMC & 26.931 & 0.8165 & 0.2317 & 0.0424 & 0.1492 & 2.419M & 315.2 & 3179.1 \\
    \MethodDPMM{} & 26.643 & 0.8144 & 0.2431 & 0.0231 & 0.0505 & \textbf{1.620M} & \textbf{384.1} & 836.2 \\
    \bottomrule
  \end{tabular}
\end{table}

Training time is reported in the last column. The posterior-tracking overhead is small: \MethodNIW{} adds 12.8~s (1.6\%) over \MethodStd{}, while \MethodDPMM{} adds 33.9~s (4.2\%) and 3DGS-MCMC requires roughly $4\times$ the training time.

We report the formal paired NIW-vs-standard comparison in the supplement. Across 39 matched scene-seed runs, the mean PSNR delta is $+0.0303$~dB with a 95\% scene-level cluster bootstrap CI of $[+0.0043, +0.0589]$~dB. A sign-flip permutation test on 13 scene-level means gives one-sided $p = 0.0327$ and two-sided $p = 0.0654$, so we interpret the result as compatibility evidence rather than as a headline rendering gain. The paired LPIPS delta is $-0.00018$ with a 95\% CI of $[-0.00038, +0.00004]$, inside a $\pm 0.001$ practical-equivalence band.

Secondary studies in the supplement bound the scope: matched PUP compression, additional active-view ladders, support-hole stress tests, the full PPU-style ranking bundle, and sparse-view or out-of-distribution (OOD)-style negative results.

\section{Limitations}
\label{sec:limitations}

The paired reconstruction effect is small. The mean PSNR delta of $+0.030$~dB has a 95\% cluster bootstrap CI that excludes zero, but the scene-level permutation test gives one-sided $p = 0.0327$ and two-sided $p = 0.0654$ with only 13 exchangeable scene units. The correct framing is therefore a small compatibility shift accompanied by useful Bayesian capability, not a large reconstruction advance.

The inference procedure is only partially closed-form. The NIW spatial update is conjugate given renderer-derived surrogate summaries, but those summaries and several secondary blocks remain optimization-driven approximations. Opacity and color receive approximate posterior-style refinements, yet hard periodic opacity reset still outperforms our Bayesian opacity alternative in all tested configurations. Tightening those approximate blocks remains open.

The uncertainty and compactness claims are intentionally scoped. Shared proxy and native geometry-induced uncertainty are different statistical objects, so the main uncertainty claim rests on native-posterior interval calibration rather than on proxy ranking metrics. The DPMM extension is a compact operating point: matched PUP remains stronger on dedicated compression and the finished OOD follow-up does not make NIW native the strongest detector, so we keep ranking-style UQ, OOD detection, and compression out of the main superiority claim set.

The evidence is also bounded in scale: 13 static scenes, 30k-iteration runs, and a single 16-to-32 active-view task family. The reported follow-ups required roughly 350 serial single-GPU hours on the hardware class described in Appendix~\ref{subsec:supp-hardware}. Larger scenes, longer horizons, dynamic settings, and planner-in-the-loop experiments may expose different approximation failures or tradeoffs.

\paragraph{Broader impacts.}
Uncertainty-aware 3D scene representations can improve the transparency and safety of confidence-aware mapping, active perception, and compact deployment. Potential negative applications include privacy-invasive reconstruction or surveillance. We use only public research benchmarks and do not introduce new person-synthesis models; any downstream release should respect the licenses and privacy expectations of upstream assets.

\section{Conclusion}
\label{sec:conclusion}

This paper presents a rendering-aware probabilistic reformulation of 3D Gaussian splatting whose native posterior improves a concrete decision task while preserving competitive quality. Three lines of evidence support this scoped claim:
\begin{enumerate}[leftmargin=1.5em,topsep=2pt,itemsep=1pt]
  \item \emph{Actionable native posterior:} in a fixed-budget 16-to-32 active-view task, native NIW acquisition improves PSNR by $+0.453$~dB over a scoring-only standard-ensemble baseline (29/39 scene-seed runs, 10/13 scene means) and by $+0.355$~dB over a protocol-matched PPU-style follow-up.
  \item \emph{Empirically calibrated intervals:} within each method's interval construction, NIW native is about $17\times$ closer to nominal 95\% coverage than the shared proxy, and about $10\times$ closer than a 3-member deep ensemble at roughly one-third the training cost.
  \item \emph{Reconstruction as a compatibility check:} a paired analysis over 39 scene-seed runs gives $+0.030$~dB PSNR (95\% CI $[{+}0.004,{+}0.059]$~dB) at only 1.6\% training overhead, supporting compatibility rather than a large leaderboard claim.
\end{enumerate}

The closed-form-versus-approximate inference boundary is part of the contribution because it makes the Bayesian claims auditable and identifies which blocks future work should tighten. The main story is deliberately narrow: active-view selection and interval calibration are positive claims, while dedicated compression and OOD-style detection remain boundary results. Taken together, the result is a 3DGS variant that is principled enough to expose explicit posterior structure over geometry and a renderer-derived inference boundary, yet practical enough to plug into the standard 3DGS rendering loop at $1.6\%$ training overhead and supply calibrated uncertainty for downstream view-acquisition decisions. Future work should test dynamic or sparse-view regimes \citep{luiten2024dynamic3dgaussians,fan2024instantsplat} and connect the posterior more directly to planner-in-the-loop tasks.

\bibliographystyle{plainnat}
\bibliography{references}

\clearpage
\appendix

\section{Extended Related Work}
\label{sec:related}

\paragraph{Neural scene representations.}
Neural radiance fields (NeRF) established volumetric neural rendering as a viable paradigm for novel-view synthesis \citep{mildenhall2020nerf}, with subsequent work improving quality and speed \citep{barron2021mipnerf,barron2022mipnerf360,muller2022instantngp}. 3DGS replaced the implicit volume with an explicit set of anisotropic Gaussians and a differentiable splatting renderer \citep{kerbl2023gaussiansplatting,zwicker2001ewa}, achieving real-time rendering with competitive quality. Our work builds on the 3DGS representation and rendering pipeline, adding Bayesian posterior structure without changing the underlying image-formation model.

\paragraph{Probabilistic reformulations of 3DGS.}
Several recent methods introduce probabilistic elements into the 3DGS pipeline. 3DGS-MCMC reframes training as stochastic-gradient Langevin dynamics with stochastic state transitions, emphasizing robustness and fixed-budget behavior \citep{kheradmand2024mcmc}. Variational Bayes Gaussian splatting (VBGS) maintains conjugate Bayesian posteriors over Gaussian parameters for continual-learning settings \citep{vandemaele2025vbgs}; however, VBGS operates on a detached density model rather than conditioning its posterior updates on renderer-derived surrogate summaries. Our approach differs from both: we keep the rendering observation model central to the posterior, track an NIW posterior over Gaussian geometry under renderer- and state-derived summaries, and couple it with a nonparametric complexity prior in a hybrid runtime whose closed-form-versus-approximate boundary is explicit.

\paragraph{Uncertainty-aware Gaussian splatting.}
Stochastic Gaussian splatting and variational multi-scale 3DGS learn probabilistic or sample-based uncertainty signals over Gaussian parameters \citep{savant2024modeling,li2024multiscaleuq}. Predictive photometric uncertainty adds a lightweight uncertainty head tied to photometric residuals \citep{galappaththige2026photometric}. These methods establish the value of uncertainty quantification for 3DGS; our distinction is to couple renderer-derived NIW geometry updates and a nonparametric complexity prior in one rendering-first probabilistic runtime, with an explicit closed-form-versus-approximate inference boundary. Unlike ensemble- or dropout-style uncertainty surrogates \citep{lakshminarayanan2017deepensembles,gal2016dropout,wilson2020bayesian}, the NIW posterior is maintained within a single training trajectory rather than requiring multiple retrains or stochastic test-time layers.

\paragraph{Pruning, sparsity, and complexity control.}
Principled uncertainty pruning (PUP) 3D-GS shows that principled pruning preserves quality better than heuristic opacity pruning under aggressive budgets \citep{hanson2025pup}. Horseshoe splatting induces structural sparsity through a hierarchical shrinkage prior on Gaussian parameters \citep{wu2026horseshoe}. Dedicated compression systems such as Compact-3DGS and LightGaussian pursue a different objective altogether: stronger end-to-end count and throughput reductions through deterministic compression and deployment-specific redesign \citep{lee2024compact3dgs,fan2024lightgaussian}. Our formulation instead tracks an explicit geometry posterior and couples it with a separate nonparametric complexity model via DP stick-breaking. We therefore view the DP layer as complementary to sparsity- or codec-oriented methods: our compression follow-up asks whether posterior summaries improve fixed-budget ranking on a matched checkpoint, not whether the current method is the strongest deployment codec. Our live comparison tables prioritize methods we reran end-to-end under one shared held-out protocol, so we keep Horseshoe splatting, predictive photometric uncertainty, and dedicated compression systems as scoped related-work references rather than direct leaderboard rows in this paper. Broader connections to Bayesian deep learning \citep{lakshminarayanan2017deepensembles,gal2016dropout,wilson2020bayesian} and variational inference \citep{jordan1999variational,blei2017variational} inform our hybrid inference design.

\section{Probabilistic Formulation}
\label{sec:supp-probabilistic}

This supplement distinguishes between the \emph{probabilistic reference model} and the \emph{current GPU trainer}. The reference model is the fully Bayesian object used to state priors, posteriors, and predictive uncertainty. The GPU trainer is the practical hybrid implementation used in the experiments.

\subsection{Reference renderer and generative model}
\label{subsec:supp-generative}

Standard 3D Gaussian splatting (3DGS) represents a scene as anisotropic Gaussians $\mathcal{G} = \{(\mu_k, \Sigma_k, o_k, c_k)\}_{k=1}^K$. Given a camera pose, each Gaussian is projected onto the image plane and the rendered color at pixel $p$ is obtained by alpha compositing in depth order:
\begin{equation}
C_p = \sum_{k \in \mathcal{N}_p} c_k \, \alpha_k \prod_{j < k} (1 - \alpha_j),
\label{eq:supp-alpha-composite}
\end{equation}
where $\alpha_k = o_k \, \mathcal{G}_{2\text{D}}(x_p \mid \mu_k^{2\text{D}}, \Sigma_k^{2\text{D}})$ is the effective contribution of Gaussian $k$ at pixel $p$ after projection, and $\mathcal{N}_p$ is the depth-sorted set of Gaussians overlapping $p$.

For Gaussian $k$, the reference model includes a normal-inverse-Wishart (NIW) geometry prior:
\begin{align}
(\mu_k, \Sigma_k) &\sim \operatorname{NIW}(m_0, \kappa_0, \nu_0, S_0), \\
o_k &\sim \operatorname{Beta}(a_0, b_0), \\
c_k &\sim \mathcal{N}(\eta_0, \tau_0^{-1} I), \\
v_k &\sim \operatorname{Beta}(1, \alpha), \qquad
\pi_k = v_k \prod_{\ell < k}(1 - v_\ell), \\
a_k &\sim \operatorname{Bernoulli}(\pi_k), \\
I_{j,p} &\sim \mathcal{N}\!\left(C_{j,p}(\Theta, a), \sigma^2 I_3\right),
\end{align}
where $C_{j,p}(\Theta, a)$ is the rendering equation under alpha-composited Gaussian splats. The NIW block gives conjugate Bayesian updates for Gaussian geometry \citep{murphy2012mlapp}; the stick-breaking construction gives a Bayesian nonparametric prior over effective component usage \citep{ferguson1973bayesian,sethuraman1994constructive,blei2006variationaldp}.

\subsection{NIW posterior updates}
\label{subsec:supp-niw-updates}

Given weighted pseudo-observations summarized by count $n$, mean $\bar{x}$, and scatter $C$, the conjugate NIW update is
\begin{align}
\kappa_n &= \kappa_0 + n, \\
m_n &= \frac{\kappa_0 m_0 + n \bar{x}}{\kappa_n}, \\
\nu_n &= \nu_0 + n, \\
S_n &= S_0 + C + \frac{\kappa_0 n}{\kappa_n}(\bar{x} - m_0)(\bar{x} - m_0)^\top.
\end{align}
The expected covariance and mean variance are then
\begin{align}
\mathbb{E}[\Sigma_k] &= \frac{S_n}{\nu_n - D - 1}, \\
\operatorname{Var}[\mu_k] &= \frac{\mathbb{E}[\Sigma_k]}{\kappa_n}.
\end{align}
In the GPU trainer, these updates are applied to rendering-derived surrogate summaries rather than exact per-pixel responsibilities. This is the main approximation that keeps million-Gaussian training tractable.

\subsection{What the current GPU trainer updates in closed form}
\label{subsec:supp-update-scope}

The live GPU trainer is a hybrid Bayesian runtime, not a literal end-to-end evidence lower bound (ELBO) optimizer. It performs explicit Bayesian posterior tracking for the spatial NIW block and the Dirichlet process (DP) stick-breaking block, while opacity and spherical-harmonic (SH) color remain anchored to the standard optimization path and receive only approximate posterior-style refinements in the default training loop even though they are part of the reference probabilistic model. The outer training render also stays on the standard 3DGS rasterizer rather than inserting activity-gated per-pixel rendering into that path; activity enters through structural control and posterior-predictive bookkeeping. This is a computational compromise: opacity and color require dense responsibility-weighted attribution under depth-ordered alpha compositing, whereas the renderer already provides efficient gradients for these parameters inside the standard 3DGS optimization path.

\section{Expanded Inference Boundary}
\label{sec:supp-boundary}

The main paper contains the compact boundary table. The fuller interpretation is:
\begin{itemize}
  \item \textbf{Closed-form in the stated submodel:} the stick-breaking update $q(v_k)$ and the residual-noise update $q(\sigma^2)$.
  \item \textbf{Conjugate under surrogate summaries:} the NIW spatial update $q(\mu_k, \Sigma_k)$, which is closed form once rendering-derived pseudo-counts and pseudo-scatter are accepted as auxiliary Gaussian statistics.
  \item \textbf{Approximate or optimization-driven:} effective component counts $\hat{N}_k$, activity gates, and the point-estimate opacity and local color refinement blocks used by the current GPU trainer.
\end{itemize}
This separation is important for claim scope. The paper does not claim exact end-to-end variational Bayes over every block at every iteration, nor does it claim activity-gated training renders in the live GPU path. It claims a rendering-aware probabilistic system whose closed-form and approximate parts are visible to the reader.

\begin{algorithm}[t]
\caption{Staged Bayesian 3DGS training. Symbolic thresholds $t_{\text{NIW}}, t_{\text{DP}}, t_{\text{post}}$ describe the family of staged runs; the canonical 30k Dirichlet process mixture model (DPMM) row instantiates them as $t_{\text{NIW}}{=}0$ (NIW geometry tracking active from the first Bayesian iteration), $t_{\text{DP}}{=}5000$ with coordinate-ascent variational inference (CAVI) every 50 iterations, and $t_{\text{post}}{=}15000$ (Table~\ref{tab:canonical_posterior_schedule}). The NIW-only row uses the same schedule with the DPMM block disabled.}
\label{alg:training}
\begin{algorithmic}[1]
\REQUIRE Scene images $\{I_j\}$, initial point cloud, NIW prior $(m_0, \kappa_0, \nu_0, S_0)$, stage thresholds $(t_{\text{NIW}}, t_{\text{DP}}, t_{\text{post}})$
\STATE Initialize $\Theta = \{(\mu_k, \Sigma_k, o_k, c_k)\}_{k=1}^K$ from the point cloud
\FOR{$t = 1, \ldots, T$}
  \STATE Render $\hat{I}_j = C_j(\Theta)$ via the standard differentiable splatting path
  \STATE Compute loss $\mathcal{L} = \|I_j - \hat{I}_j\|^2 + \lambda \mathcal{L}_{\text{SSIM}}$
  \STATE Update the Gaussian state via the standard 3DGS optimizer
  \IF{$t > t_{\text{NIW}}$}
    \STATE Extract surrogate statistics $(\hat{N}_k, \bar{x}_k, C_k)$ from the renderer
    \STATE NIW update: $q(\mu_k, \Sigma_k) \leftarrow \operatorname{NIW}(m_n, \kappa_n, \nu_n, S_n)$
    \STATE Noise update: $q(\sigma^2) \leftarrow \operatorname{IG}(\alpha_n, \beta_n)$
  \ENDIF
  \IF{$t > t_{\text{DP}}$ and DPMM enabled}
    \STATE Stick-breaking: $q(v_k) \leftarrow \operatorname{Beta}(\gamma_{k,1}, \gamma_{k,2})$
    \STATE Update activity gates $q(a_k)$; prune where $\mathbb{E}[a_k] < \epsilon$
  \ENDIF
  \IF{$t > t_{\text{post}}$}
    \STATE Apply approximate opacity-posterior refinement on a slower schedule
    \STATE Apply local linearized color-posterior refinement
  \ENDIF
  \STATE Apply standard densification and housekeeping; posterior summaries and DP activity modulate pruning decisions
\ENDFOR
\RETURN Posterior-tracked geometry with approximate auxiliary posterior refinements
\end{algorithmic}
\end{algorithm}

The publication-facing canonical schedule values are summarized in Table~\ref{tab:canonical_posterior_schedule}.

\section{Protocol and Reproducibility Details}
\label{sec:supp-protocol}

\subsection{Claim Scope Summary}
\label{subsec:supp-claim-scope}

\begin{table}[t]
  \caption{Claim-scope summary for this paper. The goal is to make explicit which statements the evidence supports, and which statements we intentionally do not make.}
  \label{tab:claim_scope_summary}
  \centering
  \scriptsize
  \setlength{\tabcolsep}{3pt}
  \begin{tabular}{p{0.25\linewidth}p{0.43\linewidth}p{0.12\linewidth}}
    \toprule
    Claim & Strongest evidence in this paper & Status \\
    \midrule
    Practical decision value & Fixed-budget active-view study: \texttt{niw\_native\_add} beats \texttt{standard\_ensemble\_add} by $+0.453$~dB over 39 runs & Yes \\
    Calibrated native posterior & NIW native coverage error $0.046$ versus $0.796$ for the shared proxy and $0.454$ for deep ensembles & Yes \\
    Near-parity reconstruction compatibility & 39-run paired NIW-vs-standard analysis with cluster bootstrap confidence interval (CI) $[{+}0.004, {+}0.059]$~dB & Yes \\
    Stronger generic uncertainty quantification (UQ) than deep ensembles & Deep ensemble area under the sparsification error (AUSE) / expected calibration error (ECE) remain better than NIW on the shared ranking-style metrics & No \\
    Stronger compression than dedicated systems & Not claimed; a matched principled uncertainty pruning (PUP) comparison shows dedicated compression stronger than our continuation runs & No \\
    Strongest out-of-distribution (OOD) detector & Full13 support-deficit OOD follow-up keeps NIW native below the proxy detector rows & No \\
    \bottomrule
  \end{tabular}
\end{table}

\subsection{Component-level evidence}
\label{subsec:supp-component-evidence}

\begin{table}[t]
  \caption{Component-level evidence for the main claims. Each row isolates the intended function of one part of the approach rather than treating all metrics as one leaderboard.}
  \label{tab:component_evidence}
  \centering
  \scriptsize
  \setlength{\tabcolsep}{3pt}
  \begin{tabular}{p{0.24\linewidth}p{0.28\linewidth}p{0.38\linewidth}}
    \toprule
    Component & Main comparison & What the result shows \\
    \midrule
    NIW geometry posterior & \MethodNIW{} vs.\ \MethodStd{}: $+0.030$~dB, $+1.6\%$ time & Posterior tracking can be added without disrupting the renderer. \\
    Native posterior samples & NIW native vs.\ NIW proxy / uniform acquisition: $+0.401$ / $+0.831$~dB & The learned posterior, not just NIW training, drives better view choices. \\
    Shared proxy UQ & NIW proxy vs.\ NIW uniform acquisition: $+0.430$~dB & Generic uncertainty helps, but native posterior uncertainty helps more. \\
    DP complexity layer & \MethodDPMM{} vs.\ \MethodStd{}: $36\%$ fewer Gaussians, $44\%$ faster frames per second (FPS), $-0.66$~dB & DP provides a compactness operating point, not the main quality claim. \\
    \bottomrule
  \end{tabular}
\end{table}

\subsection{Canonical benchmark setup}
\label{subsec:supp-canonical}

The canonical benchmark is the reported \texttt{20260405} rerun on 13 scenes from Mip-NeRF 360, Tanks and Temples, and Deep Blending, using three seeds (\texttt{6}, \texttt{66}, \texttt{666}) and five method families. The main paper uses:
\begin{itemize}
  \item \MethodStd{} as the matched-codebase deterministic control,
  \item \MethodNIW{} as the primary Bayesian row,
  \item \MethodDPMM{} as the compactness extension,
  \item \MethodOfficial{} as the external deterministic anchor,
  \item 3DGS-MCMC as the external probabilistic baseline.
\end{itemize}

Across all methods, the benchmark follows a matched evaluation policy:
\begin{itemize}
  \item one common held-out split and final-checkpoint evaluation protocol,
  \item publication-safe learned perceptual image patch similarity (LPIPS) from the vendored GraphDECO backend,
  \item common post-hoc proxy uncertainty metrics for cross-method fairness, with area under the sparsification error (AUSE) and expected calibration error (ECE) as the main fair-comparison uncertainty metrics,
  \item native posterior uncertainty reported separately for Bayesian rows, with likelihood-based calibration and coverage-error interpretation.
\end{itemize}

\begin{table}[t]
  \caption{Uncertainty evaluation design for the reported benchmark. The shared proxy and native posterior regimes answer different questions and should not be numerically mixed into one ranking.}
  \label{tab:uq_design_split}
  \centering
  \scriptsize
  \setlength{\tabcolsep}{2pt}
  \begin{tabular}{p{0.18\linewidth}p{0.16\linewidth}p{0.20\linewidth}p{0.15\linewidth}p{0.23\linewidth}}
    \toprule
    Regime & Applies to & Source of variation & Sample count & Primary use in the paper \\
    \midrule
    Shared proxy UQ & All methods & Position-jitter Monte Carlo around the final checkpoint & 50 & Fair all-method comparison via AUSE / ECE (with raw proxy coverage reported only as context) \\
    Native posterior UQ & Bayesian rows only & Posterior samples from the learned NIW / DPMM geometry posterior & 50 & Bayesian-only calibration and uncertainty-capability analysis via native negative log-likelihood (NLL), coverage error, and mean pixel / position uncertainty \\
    \bottomrule
  \end{tabular}
\end{table}

\begin{table}[t]
  \caption{Canonical staged posterior schedule used by the 30k DPMM benchmark row. The NIW-only row keeps the geometry posterior tracking path without the DP-specific activity and pruning controls.}
  \label{tab:canonical_posterior_schedule}
  \centering
  \footnotesize
  \begin{tabular}{ll}
    \toprule
    Block & Canonical setting \\
    \midrule
    Total training budget & 30k iterations \\
    NIW geometry tracking & Active from the first Bayesian training iteration \\
    DP stick-breaking CAVI & Every 50 iterations when DPMM is enabled \\
    Stage A & 0--5k: dense-count geometry / NIW / DP bookkeeping \\
    Stage B & 5k--15k: activity and residual-noise updates; count blending \\
    Stage C & 15k--30k: slower opacity and color posterior refinements \\
    Posterior-guided densification & Starts at 5k iterations \\
    Late compaction & Disabled in the canonical DPMM row \\
    \bottomrule
  \end{tabular}
\end{table}

For qualitative inspection, the analysis pipeline records the candidate render triplets, scene labels, and selection category for each inspected row before distilling the paper-facing panel. The exploratory bundle contains 11 aligned rows (4 \texttt{niw\_similar}, 4 \texttt{niw\_better}, 3 \texttt{niw\_worse}). We keep these qualitative examples as supplementary audit material rather than using them as the main paper's primary evidence.

\subsection{Dataset handling and hardware}
\label{subsec:supp-hardware}

The benchmark uses the following official-resolution image-folder policy: Mip-NeRF 360 outdoor scenes use \texttt{images\_4}, Mip-NeRF 360 indoor scenes use \texttt{images\_2}, and Tanks and Temples / Deep Blending use native \texttt{images}. The compression studies were run on a single high-memory datacenter-class GPU; the benchmark suite and follow-up studies retain the same evaluation path used for the main-paper table.

All experiments reuse existing public research assets rather than introducing a new dataset. The paper cites the original 3DGS, Mip-NeRF 360, Tanks and Temples, and Deep Blending sources, and any release package for this project should preserve the licenses and terms of use of those upstream assets as well as the external GraphDECO and 3DGS-MCMC codebases used for reported comparisons.

\begin{table}[t]
  \caption{Canonical benchmark macro summaries by dataset. These rows use the same publication-ready canonical benchmark family as the main paper table and preserve the fair shared proxy-UQ comparison through AUSE and ECE.}
  \label{tab:canonical_dataset_macro}
  \centering
  \scriptsize
  \begin{tabular}{llccccc}
    \toprule
    Dataset & Method & PSNR & SSIM & LPIPS & AUSE & ECE \\
    \midrule
    Mip-NeRF 360 & Standard 3DGS & 27.528 & 0.8160 & 0.2148 & 0.0212 & 0.0427 \\
    Mip-NeRF 360 & Bayesian 3DGS-NIW & 27.551 & 0.8161 & 0.2146 & 0.0213 & 0.0425 \\
    Mip-NeRF 360 & Official 3DGS Repro & 27.540 & 0.8162 & 0.2149 & 0.0241 & 0.0562 \\
    Mip-NeRF 360 & 3DGS-MCMC & 27.077 & 0.7905 & 0.2467 & 0.0407 & 0.1348 \\
    Mip-NeRF 360 & DPMM-3DGS & 26.871 & 0.7925 & 0.2481 & 0.0230 & 0.0468 \\
    Tanks and Temples & Standard 3DGS & 23.837 & 0.8536 & 0.1685 & 0.0289 & 0.0918 \\
    Tanks and Temples & Bayesian 3DGS-NIW & 23.904 & 0.8541 & 0.1682 & 0.0287 & 0.0925 \\
    Tanks and Temples & Official 3DGS Repro & 23.757 & 0.8533 & 0.1687 & 0.0336 & 0.1186 \\
    Tanks and Temples & 3DGS-MCMC & 23.848 & 0.8558 & 0.1546 & 0.0722 & 0.3179 \\
    Tanks and Temples & DPMM-3DGS & 23.238 & 0.8299 & 0.2037 & 0.0312 & 0.1053 \\
    Deep Blending & Standard 3DGS & 29.758 & 0.9069 & 0.2384 & 0.0141 & 0.0099 \\
    Deep Blending & Bayesian 3DGS-NIW & 29.782 & 0.9068 & 0.2384 & 0.0140 & 0.0099 \\
    Deep Blending & Official 3DGS Repro & 29.788 & 0.9076 & 0.2379 & 0.0145 & 0.0130 \\
    Deep Blending & 3DGS-MCMC & 29.355 & 0.8944 & 0.2414 & 0.0201 & 0.0458 \\
    Deep Blending & DPMM-3DGS & 29.021 & 0.8977 & 0.2603 & 0.0150 & 0.0121 \\
    \bottomrule
  \end{tabular}
\end{table}

\begin{table}[t]
  \caption{Per-scene PSNR on the canonical benchmark. The table is included to make the scene-level ranking visible without reproducing every main-table metric for every scene.}
  \label{tab:canonical_scene_psnr}
  \centering
  \scriptsize
  \begin{tabular}{llccccc}
    \toprule
    Dataset & Scene & Std. & NIW & Official & MCMC & DPMM \\
    \midrule
    Mip-NeRF 360 & bicycle & 25.26 & 25.29 & 25.26 & 25.27 & 24.56 \\
    Mip-NeRF 360 & bonsai & 32.14 & 32.24 & 32.36 & 32.09 & 31.45 \\
    Mip-NeRF 360 & counter & 29.13 & 29.13 & 29.10 & 29.10 & 28.50 \\
    Mip-NeRF 360 & flowers & 21.61 & 21.60 & 21.54 & 19.41 & 20.93 \\
    Mip-NeRF 360 & garden & 27.45 & 27.45 & 27.47 & 27.42 & 26.95 \\
    Mip-NeRF 360 & kitchen & 31.34 & 31.47 & 31.36 & 31.13 & 30.43 \\
    Mip-NeRF 360 & room & 31.66 & 31.68 & 31.59 & 31.15 & 31.09 \\
    Mip-NeRF 360 & stump & 26.62 & 26.57 & 26.62 & 26.45 & 25.58 \\
    Mip-NeRF 360 & treehill & 22.54 & 22.53 & 22.56 & 21.67 & 22.35 \\
    Tanks and Temples & train & 22.24 & 22.27 & 22.12 & 22.09 & 21.35 \\
    Tanks and Temples & truck & 25.44 & 25.54 & 25.40 & 25.60 & 25.13 \\
    Deep Blending & drjohnson & 29.33 & 29.35 & 29.41 & 28.75 & 28.47 \\
    Deep Blending & playroom & 30.19 & 30.22 & 30.17 & 29.96 & 29.58 \\
    \bottomrule
  \end{tabular}
\end{table}

\begin{table}[t]
  \caption{Paired NIW-vs-standard analysis over 39 matched scene-seed runs (13 scenes $\times$ 3 seeds). CIs use a cluster bootstrap that resamples scenes (not individual runs) to account for within-scene correlation. The exact permutation $p$-value sign-flips the 13 scene-level means. Negative LPIPS deltas favor NIW.}
  \label{tab:near_parity_stats}
  \centering
  \footnotesize
  \begin{tabular}{lcccc}
    \toprule
    Metric & Mean $\Delta$ (NIW$-$Std.) & 95\% cluster bootstrap CI & Scene wins / losses & Perm.\ $p$ (2-sided) \\
    \midrule
    PSNR & +0.0303 dB & [+0.0043, +0.0589] dB & 8 / 5 & 0.065 \\
    SSIM & +0.00011 & [-0.00018, +0.00035] & 8 / 5 & 0.484 \\
    LPIPS & -0.00018 & [-0.00038, +0.00004] & 10 / 3 & 0.131 \\
    \bottomrule
  \end{tabular}
\end{table}

Descriptively, the NIW-minus-standard macro peak signal-to-noise ratio (PSNR) delta remains positive on all three benchmark families: approximately +0.023~dB on Mip-NeRF 360, +0.067~dB on Tanks and Temples, and +0.024~dB on Deep Blending. The effect is small, but it is not concentrated in only one dataset.

\begin{figure*}[t]
  \centering
  \includegraphics[width=\textwidth]{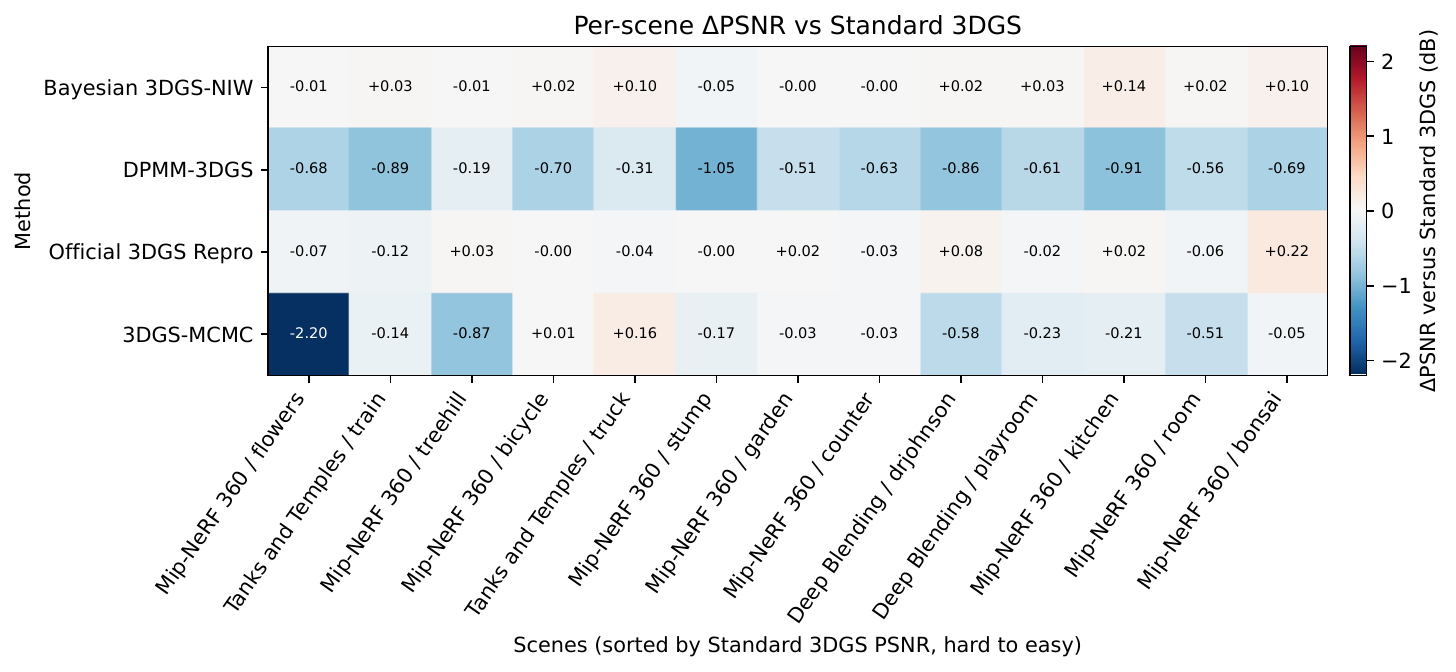}
  \caption{Per-scene $\Delta$PSNR relative to Standard 3DGS on the canonical benchmark. Columns are ordered by Standard-3DGS PSNR from hard to easy; positive values favor the row method. NIW shows small gains on several hard-to-mid difficulty scenes, while DPMM's quality cost is broad but structured.}
  \label{fig:canonical_scene_delta_psnr_heatmap}
\end{figure*}

\subsection{Deep-ensemble baseline implementation}
\label{subsec:supp-deep-ensemble}

The deep-ensemble baseline is implemented as a post-hoc replay study on top of the canonical internal-standard benchmark family rather than as a separate training code path. The goal is to compare NIW against a real uncertainty baseline without changing the held-out protocol.

\paragraph{Workflow.}
For each canonical scene, we:
\begin{enumerate}
  \item train three independent \texttt{standard\_3dgs} runs under the same 13-scene benchmark protocol using seeds \texttt{6}, \texttt{66}, and \texttt{666};
  \item export a reloadable internal checkpoint for each member from the benchmark run itself;
  \item reload the three checkpoints and render every held-out test camera;
  \item form the ensemble prediction by averaging the three red-green-blue (RGB) renders pixelwise;
  \item form the ensemble uncertainty map by taking the pixelwise standard deviation across ensemble members and then averaging over RGB channels to obtain the scalar per-pixel uncertainty consumed by the shared evaluation metrics;
  \item evaluate PSNR / structural similarity index measure (SSIM) / LPIPS on the ensemble mean render, and evaluate AUSE / ECE / negative log-likelihood (NLL) / Cov@95 on the ensemble uncertainty map.
\end{enumerate}

This baseline is therefore expensive in the standard way deep ensembles are expensive: it reuses the same final evaluation protocol, but pays roughly $3\times$ the training cost of one standard run.

\begin{table}[t]
  \caption{Deep-ensemble implementation details used for the reported UQ baseline.}
  \label{tab:deep_ensemble_impl}
  \centering
  \footnotesize
  \begin{tabular}{ll}
    \toprule
    Item & Setting \\
    \midrule
    Backbone family & \texttt{standard\_3dgs} only \\
    Ensemble size & 3 members \\
    Member seeds & \texttt{6}, \texttt{66}, \texttt{666} \\
    Data / held-out split & same canonical 13-scene benchmark split as the paper table \\
    Predictor & pixelwise mean of member RGB renders \\
    Uncertainty map & pixelwise std across members, averaged over RGB channels \\
    Quality metrics & PSNR / SSIM / LPIPS on the ensemble mean render \\
    UQ metrics & AUSE / ECE / NLL / Cov@95 on the ensemble uncertainty map \\
    \bottomrule
  \end{tabular}
\end{table}

\paragraph{Predictive photometric uncertainty (PPU)-style post-hoc comparator.}
To add a 3DGS-specific uncertainty comparator without changing the training backbone, we also ran a full13 PPU-style residual-head reimplementation on frozen \texttt{standard\_3dgs} checkpoints. The completed ranking run covers all 39 scene-seed outputs and yields macro \texttt{AUSE\_L1 = 0.0130}, \texttt{AUSE\_DSSIM = 0.0549}, \texttt{Pearson\_L1 = 0.1460}, and \texttt{Pearson\_DSSIM = 0.2300}, while leaving RGB quality unchanged because the backbone is frozen. We also evaluated a downstream \texttt{standard\_ppu\_add} policy under the same 16-to-32 active-view protocol used in the main paper. That policy improves over \texttt{standard\_uniform\_add} by $+0.547$~dB and over \texttt{standard\_ensemble\_add} by $+0.098$~dB, but still trails \texttt{niw\_native\_add} by $0.355$~dB. The main paper reports this policy as a protocol-matched sixth row in the active-view table, but it remains a separate follow-up study rather than part of the original five-policy core study. Because this study is an in-repo reimplementation and models photometric ranking rather than a calibrated predictive posterior, we keep its ranking metrics supplementary and do not mix it into the native-calibration table.

\subsection{Active-view selection implementation}
\label{subsec:supp-active-view}

The active-view selection study is the main downstream-task experiment. It tests whether native NIW uncertainty helps decide which new training views to acquire next, under a fixed final view budget and a fixed final training budget.

\paragraph{Workflow.}
For each scene-seed pair, \texttt{scripts/active\_view\_selection\_study.py} performs the following sequence:
\begin{enumerate}
  \item build the canonical train/test split using the same benchmark dataset loader and the same held-out policy (\texttt{test\_every=8});
  \item construct a deterministic 16-view seed subset from the train pool by evenly spacing indices over the original train-view order;
  \item train a phase-1 seed model on those 16 views for 10k iterations;
  \item score the remaining candidate train views according to the chosen policy;
  \item add 16 new views to reach a 32-view train set;
  \item continue training from the same model state for the remaining 20k iterations, so that every policy reaches the same final 30k-iteration budget;
  \item evaluate final PSNR / SSIM / LPIPS on the fixed held-out test split.
\end{enumerate}

\paragraph{Policy definitions.}
The core study compares five policies, and the main table additionally reports the protocol-matched PPU-style follow-up:
\begin{itemize}
  \item \textbf{Standard + uniform add}: Standard 3DGS with deterministic uniform spacing over the remaining candidate views.
  \item \textbf{Standard + ensemble-UQ add}: train three Standard 3DGS phase-1 models on the same 16-view seed subset, score candidate views by cross-member RGB variance, then continue only the first member's model for the remaining 20k iterations. This is a scoring-only ensemble acquisition rule rather than an ensemble-mean final predictor.
  \item \textbf{Standard + PPU-style add}: frozen-checkpoint PPU-style residual-head scoring on Standard 3DGS, evaluated under the same 16-to-32 acquisition protocol as a follow-up comparator.
  \item \textbf{NIW + uniform add}: NIW control with the same deterministic uniform-spacing rule.
  \item \textbf{NIW + proxy-UQ add}: NIW model, but rank candidate views by the shared proxy uncertainty score (mean proxy uncertainty over the candidate view).
  \item \textbf{NIW + native-UQ add}: NIW model, but rank candidate views by native NIW predictive uncertainty (mean native uncertainty over the candidate view).
\end{itemize}

The implementation stores the selected view indices and a preview of the highest-scoring candidate views inside every result JavaScript object notation (JSON) file so that the acquisition decisions are auditable rather than implicit.

\begin{table}[t]
  \caption{Active-view selection implementation details.}
  \label{tab:active_view_impl}
  \centering
  \footnotesize
  \begin{tabular}{p{0.27\linewidth}p{0.64\linewidth}}
    \toprule
    Item & Setting \\
    \midrule
    Initial train-view budget & 16 views \\
    Added views at acquisition step & 16 views \\
    Final train-view budget & 32 views \\
    Phase-1 training budget & 10k iterations \\
    Continuation budget & 20k iterations \\
    Final total budget & 30k iterations \\
    Compared policies & standard uniform, standard ensemble-UQ, standard PPU-style follow-up, NIW uniform, NIW proxy-UQ, NIW native-UQ \\
    Ensemble acquisition fairness & 3 phase-1 Standard models for scoring only; final predictor is a single continuation model \\
    Shared control & same deterministic 16-view seed subset per scene / seed \\
    Final metrics & PSNR / SSIM / LPIPS on the canonical held-out split \\
    \bottomrule
  \end{tabular}
\end{table}

\paragraph{Support-hole stress variant.}
We also ran a harder acquisition variant that removes a contiguous angular sector from the initial 16-view seed subset using the deterministic principal component analysis (PCA)-angle ordering already implemented in \texttt{scripts/active\_view\_selection\_study.py}. This variant keeps the same 16-to-32 acquisition budget and the same final 30k training budget as the main downstream study, but only on \texttt{flowers}, \texttt{truck}, and \texttt{playroom} across three seeds. The aim is to probe a cleaner missing-support failure mode rather than to replace the broader 13-scene result. In the raw 9-run aggregate, \texttt{niw\_native\_add} remains the strongest policy on final quality (18.7667 PSNR, 0.3631 LPIPS) versus \texttt{standard\_ensemble\_add} (18.0495, 0.3839) and \texttt{niw\_proxy\_add} (18.4826, 0.3634). However, the proxy policy averages more support-hole hits (7.89 versus 6.89 for native NIW). We therefore keep this variant as supplementary evidence for a harder downstream setting where native NIW still ends up best, but not as clean proof that the native policy most directly targets the missing sector.

\paragraph{Budget-ladder follow-ups.}
We also completed two additional active-view ladders to bound how stable the downstream advantage is away from the main 16-to-32 setting. At 24-to-48 views, \texttt{niw\_native\_add} still improves over \texttt{standard\_ensemble\_add} by $+0.2453$~dB and $-0.01194$ LPIPS, with 27/39 PSNR wins, so the positive result survives when the seed set is already larger. At 8-to-24 views, \texttt{niw\_native\_add} still beats \texttt{standard\_ensemble\_add} by $+0.7417$~dB, but no longer beats the uniform controls. We therefore cite 24-to-48 as supporting evidence and 8-to-24 as an explicit boundary-condition result rather than as a second primary downstream win.

\section{Constrained-Initialization Ablation}
\label{sec:supp-initcap}

We also tested a constrained-initialization variant in which both \MethodStd{} and \MethodNIW{} start from the same capped COLMAP seed cloud ($10\,000$ initial Gaussians) but otherwise follow the standard full $30$k training pipeline. This cap affects only the initial geometry; both methods still densify normally during training.

\begin{table}[t]
  \caption{Constrained-initialization benchmark ($10$k initial Gaussians, 13 scenes, 3 seeds, full $30$k training). NIW shows a small positive aggregate quality shift under limited initialization support.}
  \label{tab:initcap10k_macro}
  \centering
  \scriptsize
  \begin{tabular}{lccc}
    \toprule
    Method & PSNR & LPIPS & Cov@95 \\
    \midrule
    Standard 3DGS & 26.9276 & 0.2323 & 0.1684 \\
    Bayesian NIW & 26.9394 & 0.2316 & 0.1677 \\
    \midrule
    $\Delta$(NIW$-$Std.) & +0.0118 & -0.0007 & -0.0007 \\
    \bottomrule
  \end{tabular}
  \par\smallskip
  {\footnotesize PSNR wins: 19/39 scene-seed runs; positive scene-mean deltas on 8/13 scenes.}
\end{table}

\begin{table}[t]
  \caption{Per-scene mean deltas for the constrained-initialization study. Positive PSNR and negative LPIPS favor NIW.}
  \label{tab:initcap10k_scene}
  \centering
  \scriptsize
  \begin{tabular}{llccc}
    \toprule
    Dataset & Scene & $\Delta$PSNR & $\Delta$LPIPS & $\Delta$Cov@95 \\
    \midrule
    Mip360 & bicycle & -0.0100 & -0.0009 & +0.0014 \\
    Mip360 & flowers & +0.0338 & -0.0015 & -0.0021 \\
    Mip360 & garden & +0.0428 & -0.0005 & +0.0014 \\
    Mip360 & stump & +0.0325 & -0.0007 & +0.0001 \\
    Mip360 & treehill & -0.0364 & -0.0027 & -0.0007 \\
    Mip360 & room & -0.0964 & -0.0005 & -0.0020 \\
    Mip360 & counter & -0.0066 & -0.0006 & -0.0014 \\
    Mip360 & kitchen & +0.0024 & -0.0004 & -0.0043 \\
    Mip360 & bonsai & -0.0663 & -0.0001 & -0.0060 \\
    T\&T & truck & +0.0974 & -0.0006 & +0.0019 \\
    T\&T & train & +0.0113 & +0.0002 & +0.0004 \\
    DB & drjohnson & +0.0591 & -0.0003 & +0.0019 \\
    DB & playroom & +0.0900 & -0.0004 & +0.0005 \\
    \bottomrule
  \end{tabular}
\end{table}

We treat this as supporting evidence rather than a new primary claim: the effect is modest, but it survives the full 13-scene, 3-seed confirmation at the scene-mean level. The calibration story is mixed rather than uniformly positive, so this ablation supports robustness to weaker initialization more than it supports a separate uncertainty claim.

\section{Sparse-View Confirmation}
\label{sec:supp-sparse-view}

The sparse-view follow-up is a full 13-scene quick-6k stress test over train-image budgets $\{16, 32, 64, 96, 128\}$ with three seeds per budget. Each benchmark JSON stores matched \texttt{standard\_3dgs}, \texttt{bayesian\_3dgs}, and \texttt{dpmm\_3dgs} rows for the same scene-seed-budget tuple, so the macro means below are computed over 39 matched rows per budget. This study is intentionally a fast ambiguity stress test rather than a replacement for the main 30k canonical benchmark.

\begin{table}[t]
  \caption{Full13 sparse-view quick-6k macro means. NIW stays within $[-0.016,+0.023]$~dB of Standard 3DGS on macro PSNR and improves LPIPS at every budget; the scene-win columns use scene means over the three seeds.}
  \label{tab:sparse_view_confirmation}
  \centering
  \scriptsize
  \setlength{\tabcolsep}{2pt}
  \resizebox{\linewidth}{!}{%
  \begin{tabular}{rccccccccc}
    \toprule
    Budget & Std PSNR & NIW PSNR & DPMM PSNR & $\Delta$PSNR & Std LPIPS & NIW LPIPS & DPMM LPIPS & NIW PSNR wins & NIW LPIPS wins \\
    \midrule
    16  & 19.151 & 19.135 & 19.029 & -0.016 & 0.3852 & \textbf{0.3851} & 0.3886 & 5 / 13 & 7 / 13 \\
    32  & 22.173 & 22.167 & 22.116 & -0.005 & 0.3350 & \textbf{0.3347} & 0.3398 & 6 / 13 & 9 / 13 \\
    64  & 24.396 & \textbf{24.419} & 24.321 & +0.023 & 0.3088 & \textbf{0.3084} & 0.3151 & 9 / 13 & 11 / 13 \\
    96  & 24.972 & 24.970 & 24.840 & -0.002 & 0.3042 & \textbf{0.3038} & 0.3109 & 7 / 13 & 10 / 13 \\
    128 & \textbf{25.320} & 25.316 & 25.139 & -0.004 & 0.3017 & \textbf{0.3013} & 0.3086 & 8 / 13 & 12 / 13 \\
    \bottomrule
  \end{tabular}%
  }
\end{table}

\begin{table}[t]
  \caption{Full13 sparse-view compactness / throughput tradeoff. DPMM remains the compactness-oriented row across all five budgets.}
  \label{tab:sparse_view_compactness}
  \centering
  \scriptsize
  \setlength{\tabcolsep}{4pt}
  \begin{tabular}{rcccccc}
    \toprule
    Budget & Std Gauss. & NIW Gauss. & DPMM Gauss. & Std FPS & NIW FPS & DPMM FPS \\
    \midrule
    16  & 924{,}602   & 928{,}660   & \textbf{756{,}489}   & 485.1 & 482.2 & \textbf{595.3} \\
    32  & 1{,}218{,}598 & 1{,}220{,}306 & \textbf{971{,}242}   & 433.5 & 433.5 & \textbf{558.3} \\
    64  & 1{,}384{,}595 & 1{,}382{,}625 & \textbf{1{,}058{,}950} & 426.6 & 429.3 & \textbf{566.4} \\
    96  & 1{,}435{,}696 & 1{,}438{,}569 & \textbf{1{,}083{,}100} & 361.5 & 352.0 & \textbf{472.0} \\
    128 & 1{,}436{,}396 & 1{,}438{,}946 & \textbf{1{,}074{,}669} & 302.4 & 295.6 & \textbf{387.0} \\
    \bottomrule
  \end{tabular}
\end{table}

The raw recomputation confirms the intended narrow reading. NIW remains tightly clustered with Standard on PSNR across the entire budget range (macro deltas from $-0.016$ to $+0.023$~dB), but it improves LPIPS at every budget and wins more scene means on LPIPS than on PSNR. DPMM preserves its compactness-throughput operating point, using roughly 18--25\% fewer Gaussians than Standard while rendering about 23--28\% faster.

\section{Support-Deficit OOD Follow-Up}
\label{sec:supp-ood}

We also completed a bounded support-deficit OOD study. The evaluator rescans the raw \texttt{benchmark\_*.json} outputs, computes a geometric support score for each held-out test view using the normalized nearest-train-camera distance, labels the \emph{lowest-distance / best-supported} quantile as in-distribution (ID) and the \emph{highest-distance / weakest-supported} quantile as OOD, ignores the middle band, and then reports area under the receiver operating characteristic curve (AUROC), area under the precision-recall curve (AUPRC), and the mean uncertainty gap between OOD and ID views. This study probes geometric support deficit only; it is not semantic novelty or corruption robustness. The table below reports the stable full13 and hard-tail confirmations rather than the earlier exploratory run.

\begin{table}[t]
  \caption{Support-deficit OOD follow-up recomputed from raw benchmark JSONs. Each score cell reports AUROC / AUPRC. We report only the stable full13 and hard-tail confirmations. NIW native is not the strongest detector in any evaluated setting, and the stricter hard-tail split reduces the native AUROC further.}
  \label{tab:ood_support_deficit}
  \centering
  \scriptsize
  \setlength{\tabcolsep}{3.5pt}
  \begin{tabular}{lcllll}
    \toprule
    Setting & Budget & Scope & Standard proxy & Bayesian proxy & NIW native \\
    \midrule
    Full13 baseline & 6k & 13 scenes, 3 seeds & 0.570 / 0.633 & 0.567 / 0.633 & 0.557 / 0.632 \\
    Full13 baseline & 30k & 13 scenes, 3 seeds & 0.632 / 0.681 & 0.613 / 0.660 & 0.549 / 0.630 \\
    Hard-tail rerun ($q25/75$) & 30k & 13 scenes, 3 seeds & 0.604 / 0.655 & 0.634 / 0.694 & 0.545 / 0.623 \\
    Hard-tail rerun ($q15/85$) & 30k & 13 scenes, 3 seeds & 0.644 / 0.724 & 0.688 / 0.763 & 0.510 / 0.621 \\
    \bottomrule
  \end{tabular}
\end{table}

The native mean uncertainty gap is positive but small in the full13 baseline settings, and it turns slightly negative under the stricter hard-tail split ($-5.2 \times 10^{-5}$ for the NIW native row at $q15/85$). We therefore keep this study as a bounded result: it improves the empirical transparency of the paper, but it does not belong in the main claim set.

\section{Compression and DPMM Follow-Up Studies}
\label{sec:supp-compression}

Posterior-aware compression is the clearest DPMM-specific positive result. All strategies prune the same trained standard-3DGS checkpoints; the posterior strategies differ only in how the fitted DPMM posterior ranks Gaussians for retention. The reported aggregation uses all three seeds (\texttt{6}, \texttt{66}, \texttt{666}) across 39 scene-seed rows.

\paragraph{How the compaction study is implemented.}
The implementation is intentionally staged so that the comparison isolates the \emph{ranking signal} rather than conflating ranking with different training runs. For each scene-seed pair, \texttt{scripts/dp\_compression\_experiment.py} does the following:
\begin{enumerate}
  \item Train one \textbf{Standard 3DGS base model} for 30k iterations and snapshot the full Gaussian state.
  \item Load that frozen Gaussian set into a \textbf{fixed-structure DPMM posterior fit}. In the reported run we use \texttt{dp\_fit\_mode=staged\_fixed\_structure} with \texttt{dp\_inference\_steps=20000}, so the code repeatedly calls \texttt{GPUDPMMTrainer.train\_step()} but never runs the usual densify/prune loop. The Gaussian count is therefore fixed during posterior fitting; what changes are the posterior variables and the continuous Gaussian parameters updated inside \texttt{train\_step()}.
  \item Build one \textbf{score map} from that same base state and fitted posterior. The key ranking scores in the reported study are:
  \begin{itemize}
    \item \texttt{opacity}: keep high-opacity Gaussians according to the base checkpoint's sigmoid opacity values,
    \item \texttt{gradient}: keep Gaussians with large position-gradient norm on a representative training-view mean squared error (MSE) probe,
    \item \texttt{dp}: keep Gaussians with large DPMM expected mixing weights $\mathbb{E}[\pi_k]$,
    \item \texttt{dp\_posterior\_importance}: keep Gaussians with large posterior importance score $\mathbb{E}[\pi_k] \times \mathbb{E}[a_k] \times \mathbb{E}[o_k] \times \widehat{n}_k$, where $\mathbb{E}[a_k]$ is expected activity, $\mathbb{E}[o_k]$ is expected opacity, and $\widehat{n}_k$ is the normalized effective-count support term from the fitted posterior.
  \end{itemize}
  \item For each keep ratio in $\{10,20,35,55\}$ and each strategy, \textbf{clone the same base state}, prune the top-$K$ Gaussians under that score map, and then fine-tune the pruned model for 2k iterations with the standard 3DGS trainer.
  \item In the reported run, \textbf{gradient} stays single-round, but \textbf{opacity}, \textbf{dp}, and \textbf{dp\_posterior\_importance} use a two-round prune ladder (\texttt{rounds=2}). The requested total keep ratio is converted into an equal per-round keep factor. After the first prune, the code fine-tunes the pruned model, refits the DPMM posterior on that post-finetune state (\texttt{dp\_refit\_between\_rounds=true}), rebuilds the score map, and only then performs the second prune round.
  \item Finally, evaluate PSNR, SSIM, LPIPS, retained Gaussian count, and post-prune fine-tune time for every strategy-ratio pair. The reported deltas compare the primary posterior-aware strategy (\texttt{dp\_posterior\_importance}) against the opacity baseline.
\end{enumerate}

\begin{table}[t]
  \caption{Compression implementation settings used in the reported run. Every pruning strategy starts from the same Standard 3DGS base checkpoint for a given scene and seed.}
  \label{tab:compression_pipeline_details}
  \centering
  \footnotesize
  \begin{tabular}{ll}
    \toprule
    Item & Reported setting \\
    \midrule
    Base model & Standard 3DGS trained for 30k iterations \\
    Posterior fit & DPMM posterior on fixed Gaussian structure \\
    Posterior fit mode & \texttt{staged\_fixed\_structure} \\
    DP posterior-fit steps & 20k \\
    Fine-tune after pruning & 2k iterations per strategy / ratio / round \\
    Keep ratios & 10\%, 20\%, 35\%, 55\% \\
    Reported strategies & \texttt{opacity}, \texttt{gradient}, \texttt{dp}, \texttt{dp\_posterior\_importance} \\
    Primary comparison & \texttt{dp\_posterior\_importance} minus \texttt{opacity} \\
    Prune rounds & 2 for opacity / dp / posterior-aware, 1 for gradient \\
    Between-round posterior refit & Enabled \\
    \bottomrule
  \end{tabular}
\end{table}

\begin{table}[t]
  \caption{Posterior-aware compression on the 13-scene, 3-seed continuation. The posterior-aware row is strongest at 20\% keep, slightly positive again at 55\%, effectively neutral at 10\%, and slightly negative at 35\%.}
  \label{tab:compression_keep_ratios}
  \centering
  \footnotesize
  \begin{tabular}{lccccc}
    \toprule
    Keep \% & Opacity & Gradient & DP mass & Posterior & Posterior $-$ opacity \\
    \midrule
    55 & 26.049 & 25.702 & \textbf{26.124} & 26.092 & +0.042 \\
    35 & \textbf{25.898} & 24.789 & 25.793 & 25.856 & -0.042 \\
    20 & 25.403 & 22.120 & 25.373 & \textbf{25.477} & \textbf{+0.074} \\
    10 & 24.951 & 19.549 & 24.779 & \textbf{24.956} & +0.004 \\
    \bottomrule
  \end{tabular}
\end{table}

\begin{figure*}[t]
  \centering
  \includegraphics[width=\textwidth]{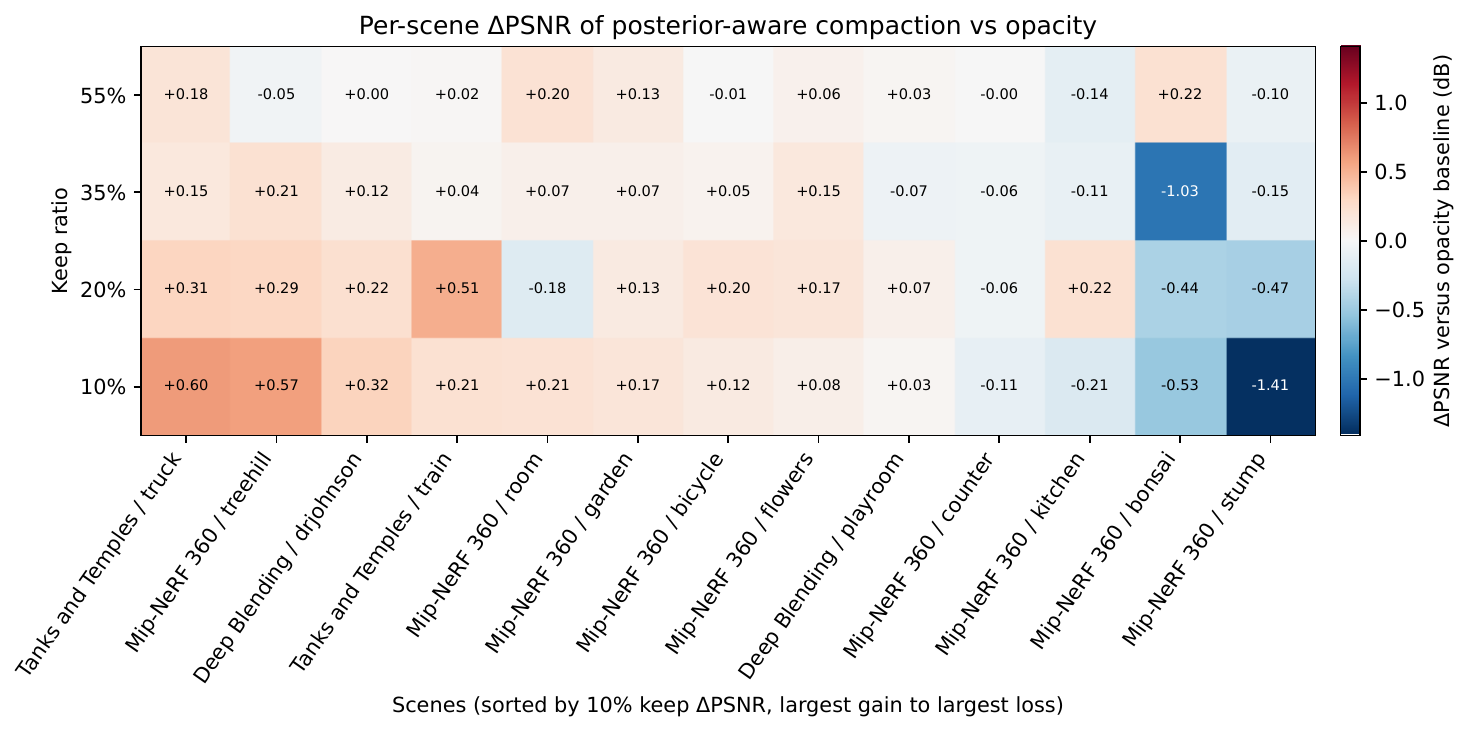}
  \caption{Per-scene $\Delta$PSNR of posterior-aware compaction (\texttt{dp\_posterior\_importance}) versus the opacity baseline across the reported 10--55\% keep ladder. Columns are sorted by the 10\% keep row from largest gain to largest loss so the aggressive-compression ranking is visually stable across the remaining ratios; positive values favor the posterior-aware ranking. The strongest positive scene means appear on \texttt{truck}, \texttt{treehill}, and \texttt{train}, while \texttt{stump} remains the clearest aggressive-budget failure case.}
  \label{fig:compression_scene_delta_psnr_heatmap}
\end{figure*}

The full three-seed recomputation changes the scene-level reading slightly relative to the earlier one-seed analysis. Posterior-aware compaction now wins on 9/13 scene means at both 10\% and 20\% keep, and on 8/13 scene means at both 35\% and 55\% keep. The largest positive scene-mean gain is on \texttt{train} at 20\% keep (+0.5084 dB), while the strongest failure is \texttt{stump} at 10\% keep (-1.4078 dB). \texttt{bonsai} remains strongly negative at the more aggressive budgets but turns positive again at 55\% keep, so the failure pattern is concentrated rather than uniformly persistent. This is why we frame the study as a meaningful but bounded Bayesian value-add for compaction with clear failure cases that still need investigation, rather than as a uniform improvement over opacity pruning. The aggregate posterior-minus-opacity deltas are +0.004 $\pm$ 0.093 at 10\% keep, +0.074 $\pm$ 0.104 at 20\%, $-0.042 \pm 0.062$ at 35\%, and +0.042 $\pm$ 0.029 at 55\%, where $\pm$ denotes the sample standard deviation across the three seed-level macro deltas.

\paragraph{Dedicated compression comparator: matched PUP.}
To bound the compression claim against a dedicated baseline, we also completed a matched two-seed PUP comparison on the same 13-scene suite. In the matched-checkpoint comparison, PUP beats the best continuation strategy at every fully covered nominal keep ratio (10/20/35/55). The clearest contrast is at nominal 20\% keep: the best continuation strategy is \texttt{dp\_posterior\_importance}, but it ends at 61.3\% of the base Gaussian count on average, whereas matched PUP ends at 20.0\% and still improves PSNR by $+1.297$~dB and LPIPS by $-0.0257$ with 26/26 PSNR wins. The separate official-checkpoint PUP macro table points in the same direction (26.693 PSNR, 0.2452 LPIPS at 20\% keep). We therefore treat the continuation study as bounded ranking evidence rather than as a claim against dedicated compression systems.

\paragraph{Default-selection follow-up.}
A full13 default-selection sweep identified \texttt{niw\_D} as the best NIW candidate by PSNR and \texttt{dpmm\_default} / \texttt{dpmm\_alpha1} as effectively tied for the best DPMM candidate within that sweep. A subsequent canonical-style confirmation rerun did not justify promoting either family: the best NIW candidate remained 0.0298~dB below the canonical NIW row, and the best DPMM candidate remained 0.6167~dB below the canonical DPMM row. We therefore keep the canonical defaults in the main paper tables.

Earlier Phase 3a and 3b mechanism studies provide appendix-level context on when posterior-aware compression is most effective:

\begin{table}[t]
  \caption{Phase 3a compression-mechanism ablation. Multi-round prune-refine helps reliably only when the DPMM posterior is refit after finetune.}
  \label{tab:phase3a_mechanism}
  \centering
  \scriptsize
  \begin{tabular}{lcccccc}
    \toprule
    Condition & Wall time & 80\% & 60\% & 40\% & 20\% & 10\% \\
    \midrule
    1round & $\sim$54 min & -0.024 & +0.003 & +0.204 & +0.214 & -0.054 \\
    2round\_refit & $\sim$5.9 h & +0.026 & -0.042 & +0.087 & +0.177 & +0.108 \\
    2round\_norefit & $\sim$91 min & +0.047 & -0.035 & +0.032 & -0.014 & +0.018 \\
    \bottomrule
  \end{tabular}
\end{table}

\begin{table}[t]
  \caption{Phase 3b factorized structural-control sweep. Late compaction is the safest single knob, but the gains are modest and remain appendix-level evidence.}
  \label{tab:phase3b_structural}
  \centering
  \footnotesize
  \begin{tabular}{lcccc}
    \toprule
    Train images & Late compaction & Split mask & Sensitivity counts & Spread \\
    \midrule
    16 & \textbf{17.0598} & 16.9221 & 16.9603 & 0.1377 \\
    32 & \textbf{20.2217} & 20.0757 & 20.0240 & 0.1977 \\
    64 & \textbf{22.6640} & 22.6152 & 22.6533 & 0.0488 \\
    \bottomrule
  \end{tabular}
\end{table}

\section{Calibration Sensitivity and Native Posterior Evidence}
\label{sec:supp-calibration}

The calibration appendix serves two purposes: it records the shared proxy-UQ results used for fair comparison, and it shows why those metrics should not be conflated with native posterior uncertainty.

\subsection{drjohnson sensitivity}
\label{subsec:supp-drjohnson}

The strongest proxy-UQ caveat is the \texttt{drjohnson} outlier from the Phase-2 contribution-isolation family:
\begin{itemize}
  \item the Phase-2 NIW proxy AUSE on \texttt{drjohnson} reached 0.2481 versus 0.0134 for standard,
  \item excluding \texttt{drjohnson} moves NIW macro AUSE from 0.0394 to 0.0221, nearly identical to standard's 0.0220,
  \item the canonical NIW \texttt{drjohnson} runs all remain in the normal 0.01357--0.01366 proxy-AUSE range.
\end{itemize}
This is why the main paper treats the outlier as a seed-sensitive proxy artifact rather than evidence that the native posterior itself is miscalibrated.

\subsection{Expanded native posterior evidence}
\label{subsec:supp-native}

The next table expands the native posterior diagnostics to scene level. It is not a fair all-method ranking because deterministic 3DGS has no native posterior counterpart. Its purpose is to show the scale and scene dependence of the Bayesian posterior outputs. We therefore treat the quantitative calibration and likelihood diagnostics in this appendix as supporting evidence for the native posterior story, with coverage error as the main comparable calibration statistic.

\begin{table}[t]
  \caption{Per-scene native posterior uncertainty diagnostics from the native-UQ evaluation path. Lower native uncertainty is not automatically better in isolation; the table is meant to show the scale and scene dependence of the native posterior outputs that deterministic 3DGS does not provide.}
  \label{tab:native_uq_per_scene}
  \centering
  \scriptsize
  \begin{tabular}{lccc|ccc}
    \toprule
    & \multicolumn{3}{c}{NIW} & \multicolumn{3}{c}{DPMM} \\
    Scene & Pix. unc. & Pos. unc. & PSNR & Pix. unc. & Pos. unc. & PSNR \\
    \midrule
    bicycle & 0.2335 & 12.69 & 25.29 & 0.0692 & 64.55 & 24.57 \\
    flowers & 0.2228 & 14.59 & 21.60 & 0.0763 & 33.05 & 20.93 \\
    garden & 0.1169 & 11.21 & 27.45 & 0.0371 & 25.02 & 26.95 \\
    stump & 0.1969 & 21.43 & 26.57 & 0.0435 & 263.32 & 25.58 \\
    treehill & 0.2381 & 13.31 & 22.53 & 0.0835 & 126.01 & 22.35 \\
    room & 0.1352 & 9.60 & 31.69 & 0.0237 & 9.96 & 31.10 \\
    counter & 0.1112 & 4.94 & 29.13 & 0.0357 & 6.47 & 28.51 \\
    kitchen & 0.1200 & 7.86 & 31.48 & 0.0339 & 14.49 & 30.43 \\
    bonsai & 0.1111 & 10.39 & 32.25 & 0.0297 & 14.37 & 31.46 \\
    truck & 0.1808 & 32.68 & 25.54 & 0.0713 & 75.05 & 25.13 \\
    train & 0.1737 & 29.73 & 22.27 & 0.0787 & 118.20 & 21.35 \\
    drjohnson & 0.0889 & 9.81 & 29.35 & 0.0254 & 9.14 & 28.47 \\
    playroom & 0.1076 & 13.81 & 30.22 & 0.0306 & 12.15 & 29.58 \\
    \bottomrule
  \end{tabular}
\end{table}

\subsection{Native uncertainty and NIW perceptual gains}
\label{subsec:supp-niw-lpips-sensitivity}

\begin{figure*}[t]
  \centering
  \includegraphics[width=\textwidth]{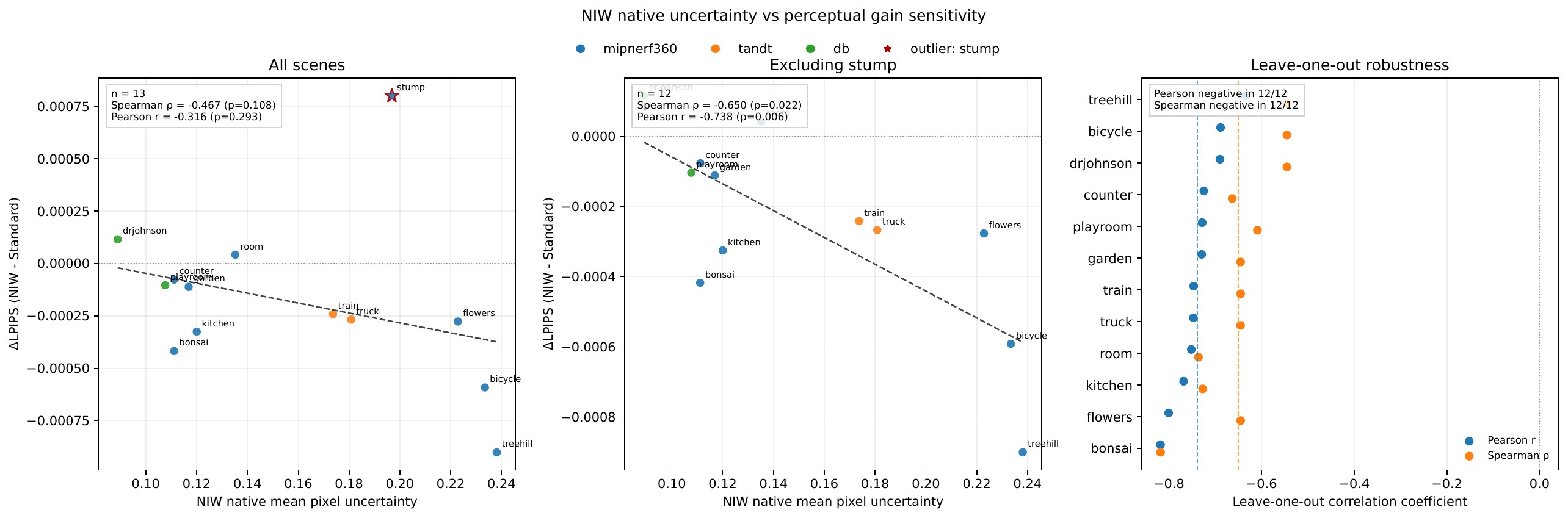}
  \caption{Scene-level sensitivity analysis for the relationship between NIW native mean pixel uncertainty and NIW LPIPS gain over Standard 3DGS. Left: all 13 canonical scenes, with \texttt{stump} highlighted as the main influence point. Middle: the same relationship after excluding \texttt{stump}. Right: leave-one-out robustness over the remaining 12 scenes, showing Pearson $r$ and Spearman $\rho$ for each omitted-scene fit.}
  \label{fig:niw_lpips_sensitivity}
\end{figure*}

We also tested whether NIW's native posterior uncertainty is informative about where NIW gains perceptual quality over the matched deterministic control. The target quantity is the scene-level LPIPS delta
\[
\Delta \mathrm{LPIPS} = \mathrm{LPIPS}_{\text{NIW}} - \mathrm{LPIPS}_{\text{Std}},
\]
so more negative values mean NIW is better. Figure~\ref{fig:niw_lpips_sensitivity} presents an outlier-transparent sensitivity analysis rather than a new primary claim.

\begin{table}[t]
  \caption{Statistical summary for the NIW native-uncertainty vs.\ LPIPS-gain sensitivity analysis. More negative $\Delta$LPIPS means a larger NIW perceptual advantage over Standard 3DGS.}
  \label{tab:niw_lpips_sensitivity_stats}
  \centering
  \scriptsize
  \begin{tabular}{lcccccc}
    \toprule
    Subset & $n$ & Spearman $\rho$ & Spearman $p$ & Spearman 95\% CI & Pearson $r$ & Pearson 95\% CI \\
    \midrule
    All scenes & 13 & -0.467 & 0.108 & [-0.938, 0.207] & -0.316 & [-0.889, 0.402] \\
    Excluding \texttt{stump} & 12 & -0.650 & 0.022 & [-0.979, -0.029] & -0.738 & [-0.928, -0.239] \\
    \bottomrule
  \end{tabular}
\end{table}

The all-scene fit is not convincing on its own: \texttt{stump} has both high native pixel uncertainty and a worse NIW LPIPS value than Standard, so it behaves as the dominant influence point. After excluding \texttt{stump}, the relationship becomes much clearer: higher NIW native pixel uncertainty is associated with more negative $\Delta$LPIPS, i.e., larger NIW perceptual gains. In that filtered analysis, Spearman reaches $\rho=-0.650$ with $p=0.022$ (permutation $p=0.026$), and Pearson reaches $r=-0.738$ with $p=0.006$ (permutation $p=0.005$).

We still treat this as supplementary evidence rather than a primary result because the exclusion is post hoc. The rightmost leave-one-out panel is included to make that caveat explicit. Even after removing \texttt{stump}, the effect is not carried by a single remaining scene: Pearson stays negative in 12/12 leave-one-out fits and remains below $p<0.05$ in all 12, while Spearman stays negative in 12/12 and remains below $p<0.05$ in 9/12. This supports the narrower claim that NIW's native pixel uncertainty is informative about where NIW improves perceptual quality, but only as an exploratory, outlier-robust sensitivity analysis.

\section{Reproducibility Notes}
\label{sec:supp-artifacts}

This preprint does not assume access to the authors' working repository. Instead, the main paper and supplement include the protocol details needed to evaluate the reported claims: dataset resolution policy, train/test scene coverage, seed policy, baseline definitions, uncertainty protocols, active-view budgets, compression keep ratios, hardware class, timing summaries, and the statistical procedures used for paired confidence intervals and sensitivity analyses.

The reported tables and figures were generated from structured run summaries for the following audit families:
\begin{itemize}
  \item canonical 13-scene, three-seed reconstruction results for Standard 3DGS, NIW, DPMM, 3DGS-MCMC, and the official implementation reference;
  \item proxy-versus-native uncertainty summaries, including per-scene interval coverage, AUSE, ECE, PSNR, SSIM, LPIPS, Gaussian count, and runtime fields;
  \item fixed-budget active-view-selection summaries for the 16-to-32 headline setting and the 8-to-24 / 24-to-48 ladder checks;
  \item compression-family summaries, including the matched-PUP comparison and scene-level delta heatmaps;
  \item qualitative render-selection metadata for the supplement panels;
  \item NIW-native uncertainty versus perceptual-gain sensitivity summaries, including the leave-one-out fits.
\end{itemize}

If the paper is accepted, the authors plan to release the scripts and structured summaries needed to regenerate the paper-facing tables and figures, subject to preserving the licenses and terms of use of the upstream public datasets and codebases.

\end{document}